\pdfoutput=1
\documentclass[10pt,twocolumn,letterpaper]{article}
\usepackage{wacv}
\usepackage{times}
\usepackage{epsfig}
\usepackage{graphicx}
\usepackage{amsmath}
\usepackage{amssymb}
\usepackage[ruled,vlined,linesnumbered]{algorithm2e}
\SetCommentSty{mycommfont}
\usepackage{graphicx}
\usepackage{subcaption}
\usepackage{mathtools}
\usepackage{comment}
\usepackage{textcomp}
\usepackage{siunitx}
\usepackage{bookmark}
\DeclarePairedDelimiter{\ceil}{\lceil}{\rceil}
\usepackage[symbol]{footmisc}

\usepackage[accsupp]{axessibility}

\let\oldnl\nl
\newcommand*{\nonl}{\renewcommand{\nl}{\let\nl\oldnl}}

\newcommand*\samethanks[1][\value{footnote}]{\footnotemark[#1]}

\hypersetup{
    colorlinks = true,
    linkbordercolor = {red},
    citecolor = {green}, 
    pagebackref = true
}

\usepackage[hyperpageref]{backref}

\begin{document}

\title{Active Learning for Improved Semi-Supervised Semantic Segmentation in Satellite Images}

\author{Shasvat Desai \thanks{Authors contributed equally}\\
Orbital Insight\\
{\tt\small shasvat.desai@orbitalinsight.com}
\and
Debasmita Ghose \samethanks\\ 
Yale University\\
{\tt\small debasmita.ghose@yale.edu}
}

%
 
\def\wacvPaperID{223} 

\wacvfinalcopy 

\ifwacvfinal
\def\assignedStartPage{} 
\fi


%
\ifwacvfinal
\else
\fi


\pagestyle{empty} 

\def\httilde{\mbox{\tt\raisebox{-.5ex}{\symbol{126}}}}

\maketitle

\begin{abstract}

Remote sensing data is crucial for applications ranging from monitoring forest fires and deforestation to tracking urbanization. Most of these tasks require dense pixel-level annotations for the model to parse visual information from limited labeled data available for these satellite images. Due to the dearth of high-quality labeled training data in this domain, there is a need to focus on semi-supervised techniques. These techniques generate pseudo-labels from a small set of labeled examples which are used to augment the labeled training set. 
This makes it necessary to have a highly representative and diverse labeled training set. Therefore, we propose to use an active learning-based sampling strategy to select a highly representative set of labeled training data. We demonstrate our proposed method's effectiveness on two existing semantic segmentation datasets containing satellite images: UC Merced Land Use Classification Dataset and DeepGlobe Land Cover Classification Dataset. We report a 27\% improvement in mIoU with as little as 2\% labeled data using active learning sampling strategies over randomly sampling the small set of labeled training data.

\end{abstract}

\begin{figure}[ht]
\centering
\begin{subfigure}{.2\textwidth} 
    \centering
    \includegraphics[width=0.55\textwidth]{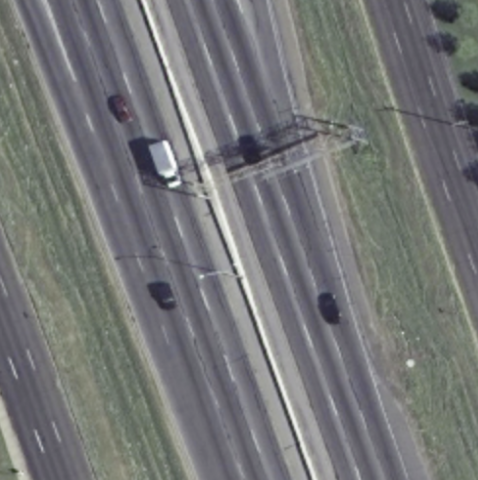}
\end{subfigure}%
\begin{subfigure}{.2\textwidth}
    \centering
    \includegraphics[width=0.55\textwidth]{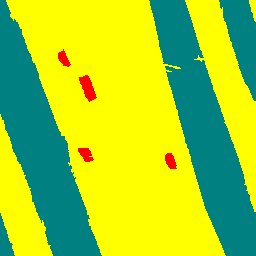}
\end{subfigure}\\
a) \hspace{80pt} b) \\
\begin{subfigure}{.2\textwidth}
    \centering
    \includegraphics[width=0.55\textwidth]{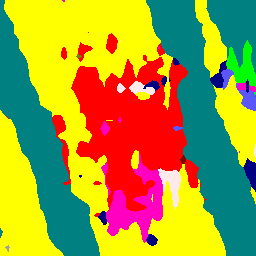}
\end{subfigure}%
\begin{subfigure}{.2\textwidth}
    \centering
    \includegraphics[width=0.55\textwidth]{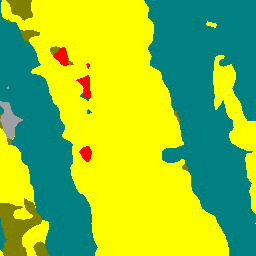}
\end{subfigure}\\
c) \hspace{80pt} d) \\
\caption{a) An image from the UC Merced Land Use Classification Dataset \cite{yang2010bag}, b) Ground Truth of the same image provided by the DLSRD dataset \cite{shao2020multilabel}, c) Baseline semi-supervised semantic segmentation model trained with 2\% labeled data, d) Output of our active learning based semi-supervised semantic segmentation model trained with 2\% labeled data. }
\label{intro-fig}
\end{figure}

 \thispagestyle{empty} 

\section{Introduction}
Semantic segmentation has found vast applications in the domain of remote sensing, including but not limited to environmental monitoring \cite{yuan2020deep, zhang2012application}, land use classification, and change detection \cite{dharani2019land, liu2017classifying, chen2020spatial, dong2020self, jianya2008review, sun2020unet}. The largest barrier to applying these segmentation techniques is the availability of representative labeled data across different geographies and terrains. Each pixel in a satellite image can represent a large area on the ground, thus requiring domain knowledge and experience to annotate pixel-level labels. This makes it significantly expensive in terms of cost and time to collect a large set of pixel-wise labels \cite{kovashka2016crowdsourcing}.  To alleviate this problem, recent work in the computer vision community has explored using fewer pixel-wise labels along with information from unlabeled images in a semi-supervised fashion \cite{hung2018adversarial, mittal2019semi, souly2017semi}. However, these small sets of images that are labeled pixel-wise are chosen randomly from a dataset \cite{hung2018adversarial, mittal2019semi}. This might bias the semi-supervised network towards a particular set of classes, degrading its performance. Therefore, we propose to use active learning to select a representative set of labeled examples for semi-supervised semantic segmentation for land cover classification.

This work is the first to explore a semi-supervised approach to semantic segmentation in satellite images to the best of our knowledge. We use a conditional GAN \cite{DBLP:journals/corr/MirzaO14} based on Mittal et al. \cite{mittal2019semi} which takes in a small number of labeled examples and a large unlabeled pool of data. This conditional GAN generates pseudo-labels based on limited labeled examples to augment the labeled pool.
This makes it essential to have a diverse set of labeled training data.  
Thus, we propose to use active learning to select a highly representative set of labeled training samples.

Active learning aims to select the most informative and representative data instances for labeling from an unlabeled data pool based on some information measure.
 We sample a subset of the images and their corresponding labels at random from a dataset, which serves as our labeled training set for the conditional GAN. We do the sampling again using an active learning-based sampling strategy which would provide a more diverse set of training data and show a performance improvement even when only very few training
samples are available. With as little as 2\% labeled data, we report an improvement of up to 27\% in mIoU over random sampling. We demonstrate our proposed method's efficacy on two existing semantic segmentation datasets containing satellite images: UC Merced Land Use Classification Dataset \cite{shao2020multilabel, yang2010bag}, and DeepGlobe Land Cover Classification Dataset \cite{demir2018deepglobe}.

Active learning for semantic segmentation \cite{mackowiak2018cereals, xie2020deal} yields patches of the given input image that are most informative. However, in this work, we require an active learning-based sampling strategy that gives us the set of most informative images from the given dataset. To achieve this, we propose using active learning for image classification to select entire images from the given dataset, which are the most informative. We then query and obtain dense-pixel level annotations only for the actively selected samples, giving us our diverse labeled training data for semi-supervised semantic segmentation.

Finally, we propose this method for sample selection to act as a guiding process for large-scale dataset creation, requiring the collection of dense pixel-level annotations. It would require significantly less cost and effort to obtain coarse image-level labels for the images and then use our proposed methodology to sample informative images labeled at pixel-level using image-classification-based active learning. The code has been made publicly available\footnote{\href{https://github.com/immuno121/ALS4GAN}{https://github.com/immuno121/ALS4GAN}}. 

Our key contributions are summarized as follows:
\begin{itemize}

    \item We use pool-based active learning sampling strategies to intelligently select labeled examples and improve performance for a GAN-based semi-supervised semantic segmentation network for satellite images.
    \item We demonstrate the applicability of the proposed method for selecting an optimal subset of data instances for which pixel-level annotations should be obtained. 
\end{itemize}

\section{Background and Related Work}

 \thispagestyle{empty} 

\subsection{Active Learning}
\label{background}
Active Learning is a technique that uses a learning algorithm which learns to select samples from an unlabeled pool of data for which the labels should be queried.
\par
\noindent
\textbf{Scenarios for Active Learning:}
Active Learning is typically employed in the following settings: \cite{settles.tr09}. \textit{Membership Query Synthesis} \cite{seung1992query} is a setting where the learner generates an instance from an underlying distribution. \textit{Stream-based Selective Sampling} \cite{atlas1990training} queries each unlabeled instance one at a time based on some information measure. The last scenario  \textit{Pool-based sampling}, used in this paper,  assumes a large pool of unlabeled data and draws instances from the pool according to some information measure.
\par
\noindent
\textbf{Query Strategies:}
There are several query strategies in the pool-based sampling strategies to select samples for which we need to query labels. 
\par
The \textit{margin sampling strategy} \cite{balcan2007margin, roth2006margin} selects the instance that has the smallest difference between the first and second most probable labels.
\begin{align}
   x_{M} &= \underset{x}{\arg\max} [P_{\theta} ({\hat{y_2}}|x) - P_{\theta} ({\hat{y_1}}|x)]
\end{align}
Intuitively, instances with small margins are more ambiguous and knowing the true label should help the model discriminate more effectively.
 \par
The third and the most common strategy is \textit{Entropy-based sampling} \cite{lewis1994heterogeneous, lewis1994sequential}. 
\begin{align}
    x_{H}^{*} 
    &= \underset{x}{\arg\max}  - \sum{_y}P_{\theta}(y|x)\log(P_{\theta}(y|x))
\end{align}
where $y$ ranges over all possible labels of $x$. Entropy is a measure of a variable’s average information
content. So intuitively, this method selects samples by ranking them based on their information content. 

\noindent
\textbf{Applications: } Active learning techniques have found numerous applications in medical imaging \cite{hoi2006batch, joshi2012scalable,mahapatra2018efficient, shao2018deep, smailagic2018medal} and remote sensing \cite{kellenberger2019half, lin2020active, rodriguez2021mapping, stumpf2013active, tuia2009active, tuia2011survey}  communities because obtaining labeled data in those domains have been particularly challenging \cite{paisitkriangkrai2016semantic, SCHLEMPER2019197}.  
Some recent work have used deep learning techniques for active learning-based image classification \cite{ranganathan2017deep, shao2018deep}, semantic segmentation \cite{mackowiak2018cereals,xie2020deal,yang2017suggestive} and object detection \cite{roy2018deep}. However, to the best of our knowledge, this is the first work that uses a deep active learning-based image classifier to select labeled examples for a semi-supervised semantic segmentation network.

\subsection{Semi-Supervised Semantic Segmentation}
 \thispagestyle{empty} 
Most of the existing semi-supervised semantic segmentation techniques either use  consistency regularization strategies \cite{chen2021mask, french2020semi, olsson2020consistency, olsson2020classmix, ouali2020semi} or generative model to augment pseudo-labels to existing labeled data pool \cite{hung2018adversarial, souly2017semi} or some combination of the two \cite{mittal2019semi}.  

\noindent
\textbf{Consistency Regularization-based strategies:} The core idea in consistency regularization is that predictions for unlabeled data should be invariant to perturbations. 
Some recent work has pointed out the difficulty in performing consistency regularization for semi-supervised semantic segmentation because it violates the cluster assumption \cite{french2020semi, ouali2020semi}. 
Some other work
\cite{chen2021semisupervised, french2020semi, kim2020structured} use data augmentation techniques like
CutMix \cite{yun2019cutmix} and ClassMix \cite{olsson2020classmix}, which composite new images by mixing two original images. They hypothesize that this would enforce consistency over highly varied mixed samples while respecting the original images' semantic boundaries.

\begin{figure*}[ht]
    \centering
    \includegraphics[scale=0.29]{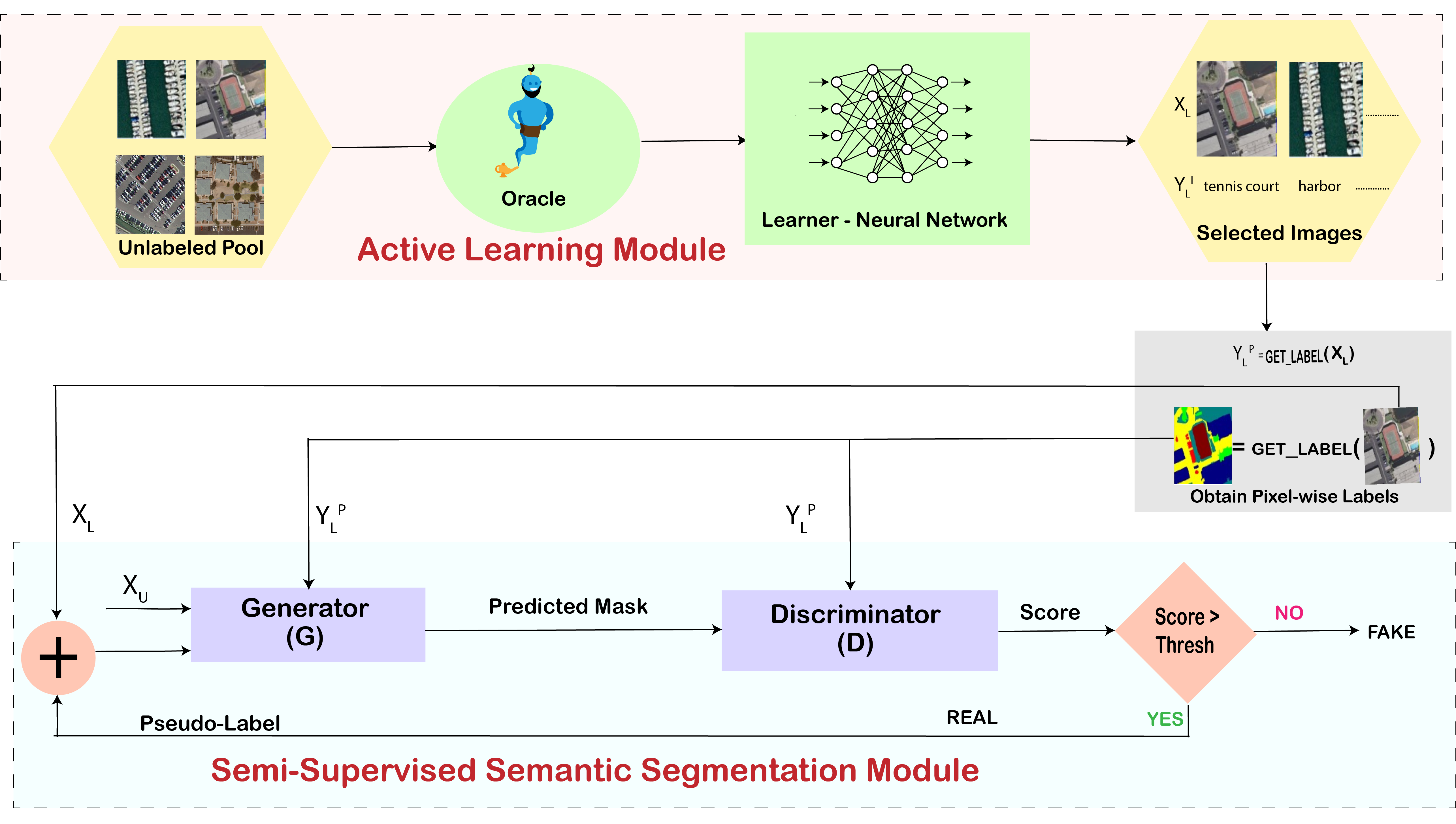}
    \caption{\textbf{Proposed Framework:} 1) The active learning module expects an unlabeled pool of data as its input. It is an image classification network that returns samples $X_L$ selected based on some information measure determined by the sampling strategy used for the active learner.  2) A \textit{get\_label} operation is performed to obtain pixel-level labels corresponding to images returned by the active learning module. 3) A conditional GAN is then trained where the generator module is a semantic segmentation network and expects the labeled images returned by the active learning module, $X_{L}$, the corresponding  pixel-wise labels for these samples $Y_{L}^P$, along with the remaining unlabeled images  $X_{U}$. It outputs a segmentation mask. 4) The discriminator expects the predicted segmentation masks from the generator along with the pixel-wise ground truth labels  $Y_{L}^P$, and outputs a prediction confidence score. 5) prediction masks with a score greater than the predefined confidence threshold $\tau$ are selected and treated as pseudo-labels to train the GAN and are augmented to the labeled pool as shown by the \textit{"+"} sign on the bottom left corner.
    }
    \label{fig:framework}
\end{figure*}
\noindent
\textbf{GAN-based strategies:} Souly et al. \cite{souly2017semi} was the first work to perform semi-supervised semantic segmentation using a GAN. They employ the generator to generate realistic visual data that forced the discriminator to learn better features for more accurate pixel classification. 
However, these generated images were not sufficiently close to the real images since it is challenging to generate realistic-looking images from pixel-wise maps. 
\par
To overcome the drawback of poorly generated images, Hung et al. \cite{hung2018adversarial} propose a conditional GAN. The generator is a standard semantic segmentation network that takes in images and their ground truth maps. The discriminator is a fully convolutional network (FCN) that takes the ground truth, and the segmentation map predicted by the generator and aims to distinguish between the two.
Thus, it is difficult for the discriminator to determine if the pixels belong to the real or the fake distribution by looking at one pixel at a time without context. 
\par
Mittal et al. \cite{mittal2019semi} propose to replace the FCN-based discriminator with an image-wise discriminator that determines if the image belongs to the real or the fake distribution, which is a relatively easy task. Additionally, they propose to use a supervised multi-label classification branch \cite{tarvainen2017mean} which decides on the classes present in the image and thus aids the segmentation network to make globally consistent decisions. During evaluation, they fuse the two branches to alleviate both low-level and high-level artifacts that often occur when working in a low-data regime.
In this work, we use the s4GAN branch of the network presented by Mittal et al. \cite{mittal2019semi} and propose to select a more optimal set of labeled examples to improve the performance of the network over a random selection of labeled examples.

\section{Our Method}
We propose to use active learning techniques to select a small informative subset of labeled data that would help the semi-supervised semantic segmentation model learn more effectively with a representative pool of labeled data.  The proposed framework is demonstrated in Figure \ref{fig:framework}. It should be noted that for any given image, our method assumes the ability to gain access to its corresponding image and pixel-level annotations. 

\begin{algorithm}[t]
\SetAlgoLined
\nonl {\textbf{Input:}} \\
Labeled ratio $R$ and unlabeled pool of data $X_N$

\nonl {\textbf{Output:}} \\
Informative samples and their image-level labels ($X_{L}$,$Y_L^I$)

\nonl {\textbf{Define:}}
 $learner \leftarrow \text{Neural network based image classifier} $\label{learner} \\
 $oracle \leftarrow \text{source of labels} $ \label{oracle}\\
 Number of data points to sample: $X_{N_L} \leftarrow R * X_N$\label{labeled_samples} \\
 Size of initial labeled pool for learner: $init\_size \leftarrow \ceil{\alpha_{init} * X_{N_L}} $\label{init_size} \\
 Initial labeled pool for active learner:($X_{init}$, $Y_{init}$)\label{init_pool}\\
Train the active learner:
 $learner(X_{init}, Y_{init})$ \label{learner_teach_step} \\
 Unlabeled pool: $X_{pool} \leftarrow X_N - X_{init} $ \label{X_pool}\\
 $X_L \leftarrow \{\}$, $Y_L^I \leftarrow \{\}$ \label{init_vars}\\
 Number of samples to query in each iteration : $N_Q = \beta_Q * init\_size$ \label{X_Q}\\
\nonl \While{$n(X_L)\leq X_{N_L}$}{
  {\nonl \textbf{Query Step:}} \\
  \nonl Inference on unlabeled pool using $learner$: \\
  $prediction\_scores = learner(X_{pool})$  \label{pred_scores}\\
  \nonl Select top-Q most informative instances ($N_Q$): \\
  $X_Q$ = $sampling\_strategy(prediction\_scores)$ \\
  $Y_Q = oracle(X_{Q})$ \label{oracle_query_step}\\
  $X_{pool} = X_{pool} - X_Q$ \label{pool_remove_Xl}\\
  $X_L = X_L \bigcup X_Q$ \label{X_union}\\
  $Y_L^I = Y_L^I\bigcup Y_Q$ \label{Y_union}\\
  \nonl {\textbf{Teach Step:}} \\
  \nonl Retrain the $learner$ with updated labeled pool: \\
  $learner(X_L, Y_L^I)$ \label{learner_teach2_step} 
 \nonl }
 \Return $X_L, Y_L^I$
 \caption{Active Learning for Labeled Sample Selection}
 \label{AL}
\end{algorithm}

\subsection{Active Learning for Image Classification}
 Algorithm \ref{AL} describes how active learning was used to select the most informative and diverse set of labeled samples for semi-supervised semantic segmentation. 
We use active image classification and sample images using pool-based sampling strategies \cite{settles.tr09} as described in Section \ref{background}. 
Algorithm \ref{AL} is used as an offline process to sample informative samples and it accepts two inputs: labeled ratio $R$, and an unlabeled pool of data, $X_N$. The labeled ratio $R$, determines the number of labeled samples used to train the semi-supervised model. The labeled ratio $R$, used in this paper for each dataset, can be found in Table \ref{table_1}. 
\par
\noindent
\textbf{Initialization:}
The active learner is initialized with $init\_size$ number of image-level labels (line \ref{init_pool}), which is a function of the number of labeled samples to be returned. We define the parameter $\alpha_{init} \in (0, 1]$ (line \ref{init_size}) to control the size of the initial labeled pool of the active learner. $\alpha_{init}$ helps in determining the optimal size of the labeled pool that the active learner should be initialized with for every labeled ratio, $R$. A low value of $\alpha_{init}$ would result in the active learner being initialized with a tiny pool of data not providing sufficient information about the data distribution. In contrast, a large value of $\alpha_{init}$  might bias the active learner toward a particular set of initial samples, which might lead to under-sampling of a particular class. Intuitively, setting $\alpha_{init}$ to 1 would make to outcome close to being equivalent to random sampling. We found this approach to perform better than having a fixed initial labeled pool size, irrespective of the labeled ratio, $R$. The $learner$ is then trained with this initial labeled pool (line \ref{learner_teach_step}). Once the initial labeled samples are selected for the active learner, they are removed from the unlabeled pool (line \ref{X_pool}).
 We sample data instances and their labels by performing the query and teach steps in an interleaved fashion.
 
\noindent
\textbf{Query Step:}
 We run inference using the trained $learner$ on the entire unlabeled pool (line \ref{pred_scores}) and obtain the model's confidence scores for each sample in that pool. Then using some uncertainty measure based on the active learning strategy used, the oracle queries image-level labels for top-Q uncertain samples, $N_Q$. For instance, if entropy-based sampling \cite{lewis1994heterogeneous, lewis1994sequential} is used, then the oracle will return labels for samples with the highest entropy measure.  The optimal number of data instances $N_Q$, queried from the $oracle$ in every iteration is a function of the initial labeled pool size $init\_size$ and another parameter, $\beta_Q \in (0, 1]$ (line \ref{X_Q}). We define $\beta_Q$ to determine the number of iterations for which the active learner will be trained. It is crucial for the active learner's performance because a small value of $\beta_Q$ will add only a small number of labels in each iteration, resulting in a negligible weight update of the active learner. In contrast, a large value of $\beta_Q$ will cause a massive update in the learner's weights at every step. It will also reduce the total number of steps the learning algorithm will take to reach its target ${X_{N_L}}$. This will leave little room for the learner to learn from its mistakes in each iteration, directly impacting the quality labels produced.
Once the labels for $N_Q$ number of samples are queried from $oracle$, the images and their corresponding labels are added to the result set, $X_L$ and $Y_L^I$ (lines \ref{X_union},  \ref{Y_union}).

 \thispagestyle{empty} 

\noindent
\textbf{Teach Step:}
 In this step, the $learner$ is trained with the updated labeled pool of samples obtained from the query step (line \ref{learner_teach2_step}).
 The image classification network's capacity for the $learner$ is also crucial in determining the quality of the selected samples. Any network with low capacity tends to underfit, while any network with a higher capacity than required could overfit and detrimentally affect the downstream task's performance. 

\begin{table}[] 
    \centering
    \begin{tabular}{p{0.35\linewidth} p{0.1\linewidth} p{0.1\linewidth} p{0.1\linewidth} p{0.1\linewidth}} 
    \hline
      \textbf{Labeled Ratio(R)}  &  \textbf{2\%} & \textbf{5\%} & \textbf{12.5\%} & \textbf{100\%} \\
      \hline
       UC Merced\cite{yang2010bag} & 34 & 85 & 211 & 1680 \\
       
       DeepGlobe\cite{demir2018deepglobe}& 12 & 32 & 80 & 642\\
       \hline
    \end{tabular}
    \caption{Number of Labeled Examples per Labeled Ratio in the UC Merced and DeepGlobe datasets}
    \label{data-table}
\label{table_1}
\end{table}

\subsection{Semi-Supervised Semantic Segmentation}

\label{cond-GAN}

We use the s4GAN network proposed by Mittal et al. \cite{mittal2019semi} for performing semi-supervised semantic segmentation using a small number of pixel-wise labeled examples along with a pool of unlabeled examples. This is a conditional GAN-based technique where the generator $G$ is a segmentation network. The generator takes in all the labeled and unlabeled images, along with the ground truth masks.
The discriminator $D$ takes the predicted segmentation map and the available ground truth masks concatenated with their respective images. The network attempts to match the real and the predicted segmentation maps' distribution through adversarial training.

\noindent
\textbf{Notation:} \newline
$x_L^P, y_L^P$: image with their pixel-wise labels \newline
$x_U^P$:  image with no pixel-wise ground-truth labels

\subsubsection{Segmentation Network (Generator)}

\label{seg-network}

 \thispagestyle{empty} 
The segmentation network S is trained with loss $L_S$, which
is a combination of three losses: the standard cross-entropy
loss, the feature matching loss, and the self-training loss.
\noindent
\textbf{Cross Entropy Loss:}  Standard supervised pixel-wise cross entropy loss term evaluated only for the labeled samples $x_L^p$ is shown is Equation \ref{ce loss}.
\begin{equation}
\label{ce loss}
L_{ce} = -\sum y_L^P \cdot log(S(x_L^P))
\end{equation}
\newline
\textbf{Feature-Matching Loss: } The feature matching loss $L_{fm}$ \cite{salimans2016improved} aims to minimize the mean discrepancy between the feature statistics of the predicted, $S(x_U^P)$ and the ground truth segmentation maps, $y_L^P$ as shown in Equation \ref{fm loss}. This loss uses both labeled and unlabeled training examples. 
\begin{multline}
    \label{fm loss}
    L_{fm} = || \mathbb{E}_{(y_L^P, x_L^P) \sim D_l} D(x_L^P \oplus y_L^P)  - \\
    \mathbb{E}_{(x_U^P) \sim D_u} D(x_U^P \oplus S(x_U^P))|| 
\end{multline}
\newline
\textbf{Self-Training Loss: } This loss is used for only unlabeled data. This loss aims to pick the best outputs of the segmentation network (i.e., those outputs that could fool the discriminator) that do not have a corresponding ground truth mask and reuse them for supervised training.
Intuitively, it pushes the segmentation network to produce predictions that the discriminator cannot distinguish from real. The discriminator's output is a score between 0 and 1, denoting the discriminator's confidence that the predicted segmentation mask is real. The predicted segmentation mask with a score greater than the predefined confidence threshold $\tau$, is selected and treated as a pseudo-label to train the GAN.

Equation \ref{st loss} describes the self-training loss. 
\begin{equation}
\label{st loss}
    L_{st} = 
    \begin{cases}
   -\sum{\textbf{y*} \cdot log(S(x_U^P))} & \text{if } D(x_U^P) \geq \tau \\
   0 & \text{otherwise}
    \end{cases}
\end{equation}
$\textbf{y*}$  = pseudo pixel-wise labels which are the predictions of the segmentation network 
\newline
Finally, the objective function for the generator is given by Equation \ref{gen loss}.  
\begin{equation}
\label{gen loss}
    L_S = L_{ce} + \lambda_{fm}L_{fm} + \lambda_{st}L_{st}
\end{equation}
where, $\lambda_{fm} > 0$ and $\lambda_{st} > 0$ are the weighting parameters for the feature matching and the self-training losses.

\subsubsection{Discriminator}

The discriminator is trained to distinguish between the real labeled examples and the fake segmentation masks generated by the network concatenated with the corresponding input images. It is trained using the original GAN loss proposed by Goodfellow et al. \cite{goodfellow2014generative} as shown in Equation \ref{disc loss}. 

\begin{multline}
    \label{disc loss}
        L_{D} = \mathbb{E}_{(y_L^P, x_L^P) \sim D_l} log D(x_L^P \oplus y_L^P)  + \\
   \mathbb{E}_{(x_U^P) \sim D_u} log(1-D(x_U^P \oplus S(x_U^P))
\end{multline}

\subsection{Labeled Example Selection for Semi-Supervised Semantic Segmentation using Active Learning}
To obtain labeled samples for a semi-supervised semantic segmentation network, the proposed framework uses the active learning module from Algorithm \ref{AL} defined as $Active\_Sampler$ in Algorithm \ref{AL_2} and shown in Figure \ref{fig:framework}. The active learning module expects an unlabeled pool of data ($X_N$) and the labeled ratio $R$ as its input. It returns samples $X_{N_L}$ (line \ref{sampler}), which are selected based on some information measure determined by the sampling strategy used for the active learner. The active learning module is called only once to select informative data instances for the semi-supervised model.
In the get\_label stage, we obtain pixel-level labels corresponding to only those images returned by the $Active\_Sampler$ to serve as the initial labeled training data for the semi-supervised segmentation module. This enables the semi-supervised semantic segmentation network to learn with a representative labeled set instead of a random subset of the data.
\par
The conditional GAN described in Section \ref{cond-GAN} is then trained, where the generator module is a semantic segmentation network and expects the labeled images returned by the active learning module $X_{N_L}$, the corresponding pixel-wise labels for these samples $Y_{N_L}^P$, along with the remaining unlabeled images, $X_{N_U}$. The output of the generator network is a segmentation mask. The discriminator expects the predicted masks along with the pixel-wise ground truth labels  $Y_{N_L}^P$, and outputs a probability score between 0 and 1,  denoting its confidence in the predicted mask being real and belonging to the ground truth. 
If this confidence score is greater than the predefined confidence threshold $\tau$, then it implies the generator has successfully predicted a mask that appears real to the discriminator. Hence, this prediction is augmented to the ground-truth, $Y_{N_L}^P$ (line \ref{union}) and used as a pseudo-label, and the GAN is trained with this updated dataset. These pseudo-labels contribute to the self-training loss detailed in Section \ref{seg-network}.  The generator and the discriminator are trained adversarially until a stopping criterion is satisfied. 

\begin{figure*}[ht]
\centering
2\% \hspace{8pt}
\begin{subfigure}{.2\linewidth}
    \centering
    \includegraphics[width=0.7\linewidth]{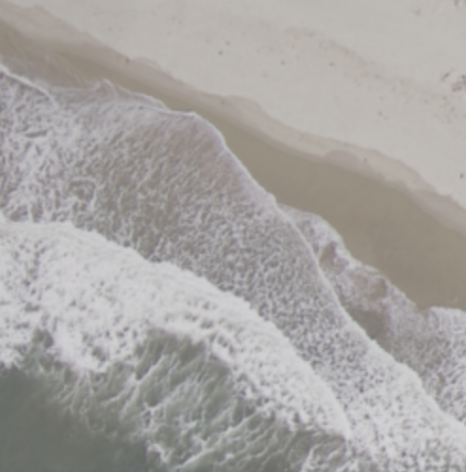}
\end{subfigure}%
\begin{subfigure}{.2\linewidth}
    \centering
    \includegraphics[width=0.7\linewidth]{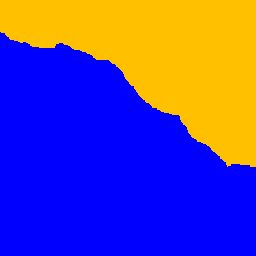}
\end{subfigure}
\begin{subfigure}{.2\linewidth}
    \centering
    \includegraphics[width=0.7\linewidth]{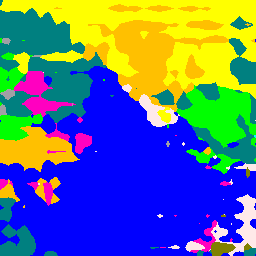}
\end{subfigure}%
\begin{subfigure}{.2\linewidth}
    \centering
    \includegraphics[width=0.7\linewidth]{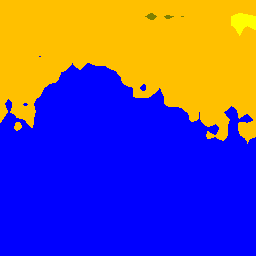}
\end{subfigure}\\
5\% \hspace{8pt}
\begin{subfigure}{.2\linewidth}
    \centering
    \includegraphics[width=0.7\linewidth]{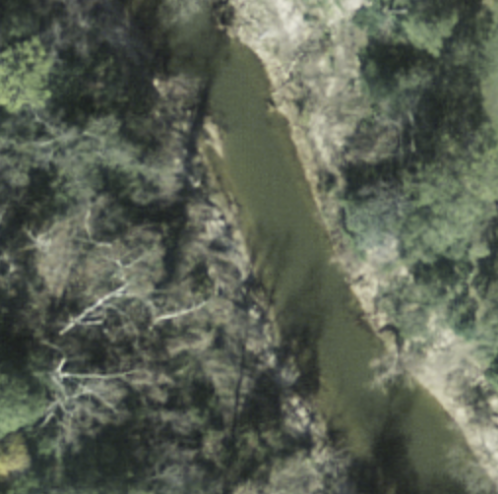}
\end{subfigure}%
\begin{subfigure}{.2\linewidth}
    \centering
    \includegraphics[width=0.7\linewidth]{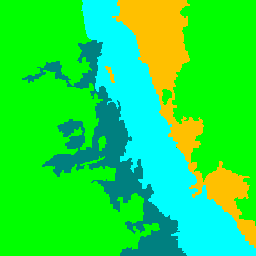}
\end{subfigure}
\begin{subfigure}{.2\linewidth}
    \centering
    \includegraphics[width=0.7\linewidth]{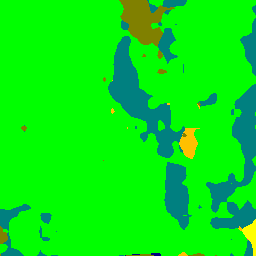}
\end{subfigure}%
\begin{subfigure}{.2\linewidth}
    \centering
    \includegraphics[width=0.7\linewidth]{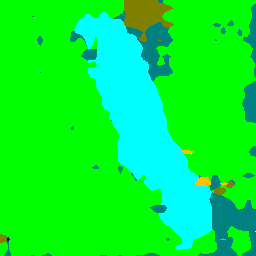}
\end{subfigure}\\
12.5\% 
\begin{subfigure}{.2\linewidth}
    \centering
    \includegraphics[width=0.7\linewidth]{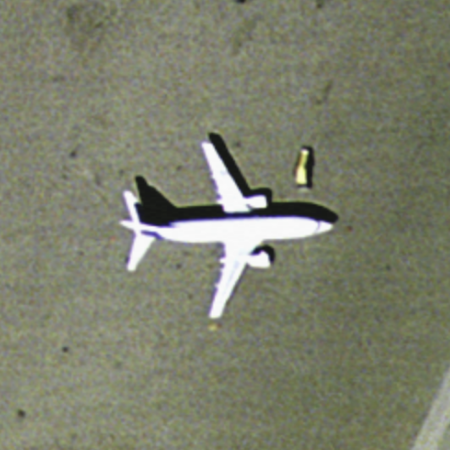}
\end{subfigure}%
\begin{subfigure}{.2\linewidth}
    \centering
    \includegraphics[width=0.7\linewidth]{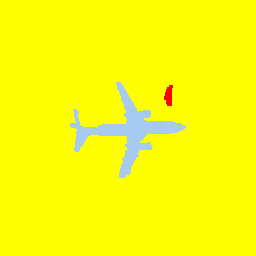}
\end{subfigure}
\begin{subfigure}{.2\linewidth}
    \centering
    \includegraphics[width=0.7\linewidth]{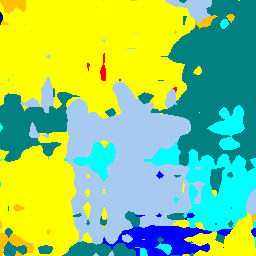}
\end{subfigure}%
\begin{subfigure}{.2\linewidth}
    \centering
    \includegraphics[width=0.7\linewidth]{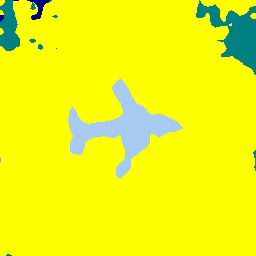}
\end{subfigure}\\
\begin{center}
    \hspace{5pt} a) \textbf{Original Image} \hspace{5pt} b) \textbf{Ground Truth} \hspace{30pt} c) \textbf{Baseline} \hspace{35pt} d) \textbf{Our Results}
\end{center}
\caption{Qualitative Results from the UC Merced Land Use Classification Dataset for different labeled ratios}
\label{qualtitative-ucm}
\end{figure*}

\begin{figure*}[ht]
\centering
2\% \hspace{8pt}
\begin{subfigure}{.2\linewidth}
    \centering
    \includegraphics[width=0.7\linewidth]{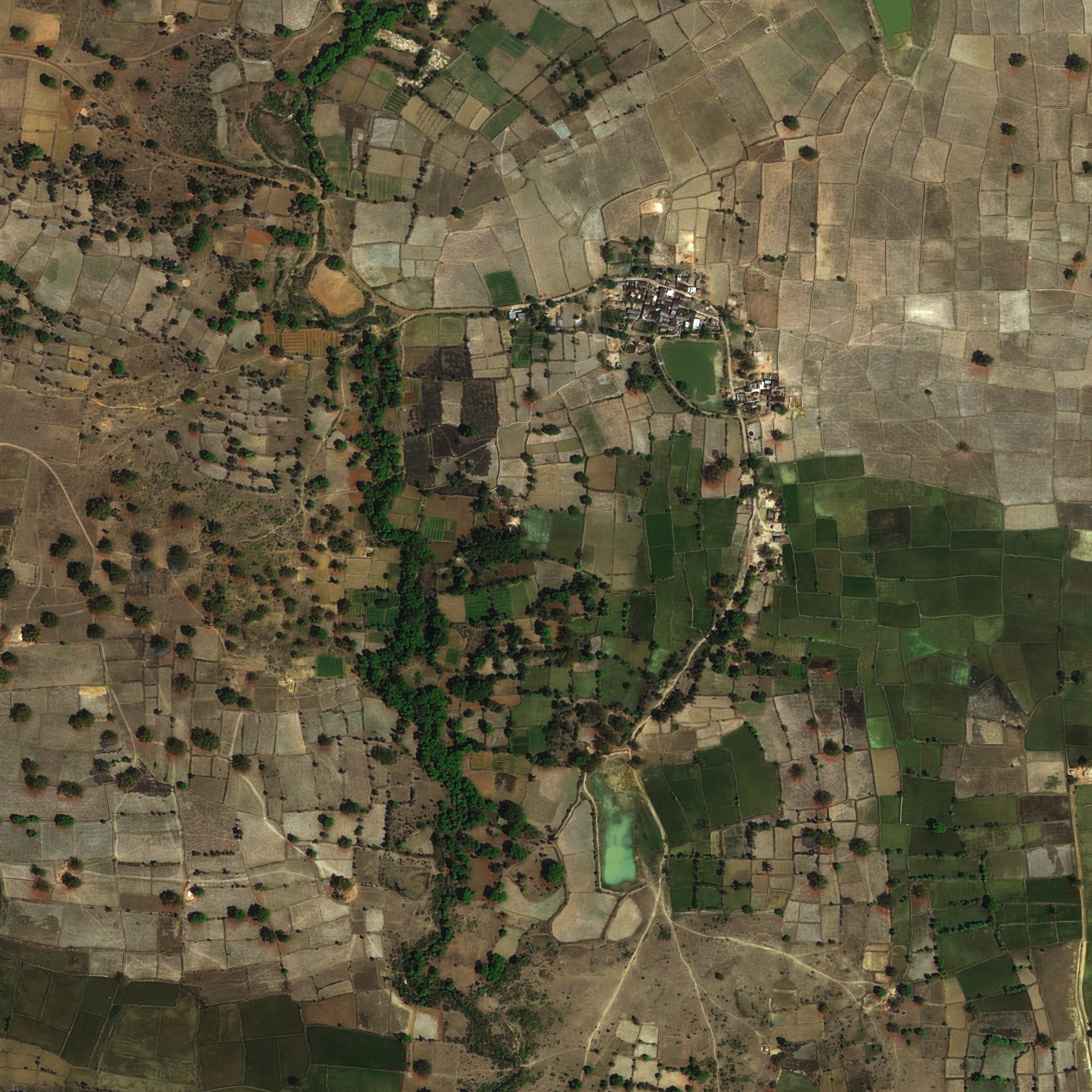}
\end{subfigure}%
\begin{subfigure}{.2\linewidth}
    \centering
    \includegraphics[width=0.7\linewidth]{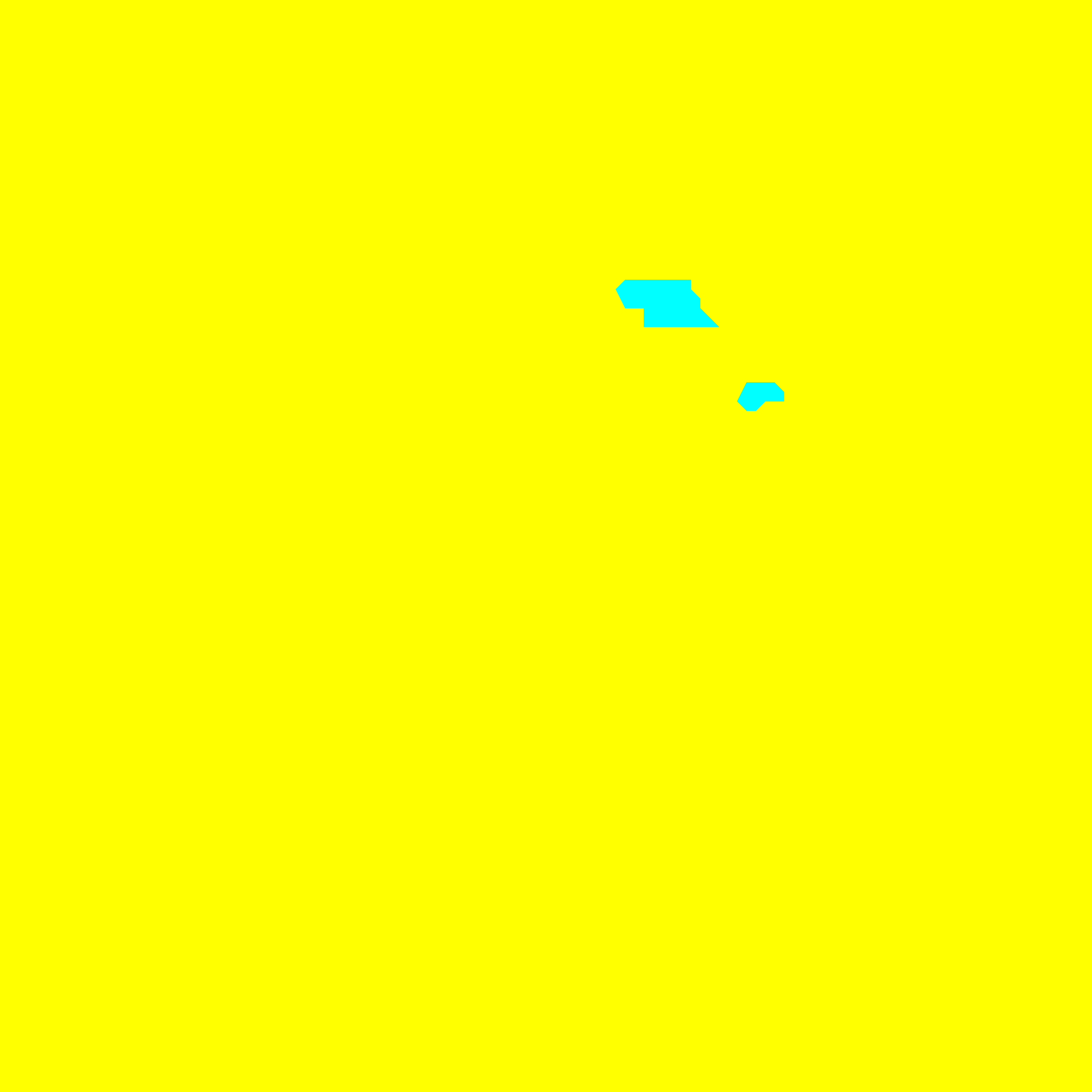}
\end{subfigure}
\begin{subfigure}{.2\linewidth}
    \centering
    \includegraphics[width=0.7\linewidth]{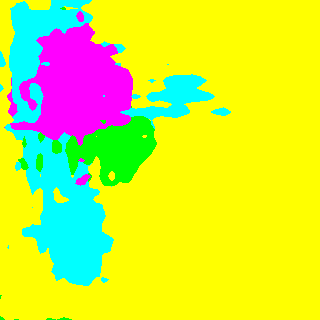}
\end{subfigure}%
\begin{subfigure}{.2\linewidth}
    \centering
    \includegraphics[width=0.7\linewidth]{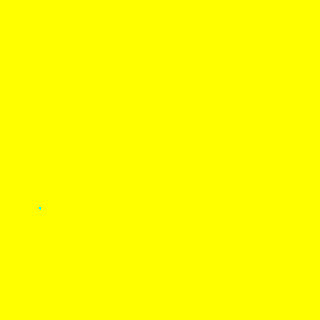}
\end{subfigure}\\
5\% \hspace{8pt}
\begin{subfigure}{.2\linewidth}
    \centering
    \includegraphics[width=0.7\linewidth]{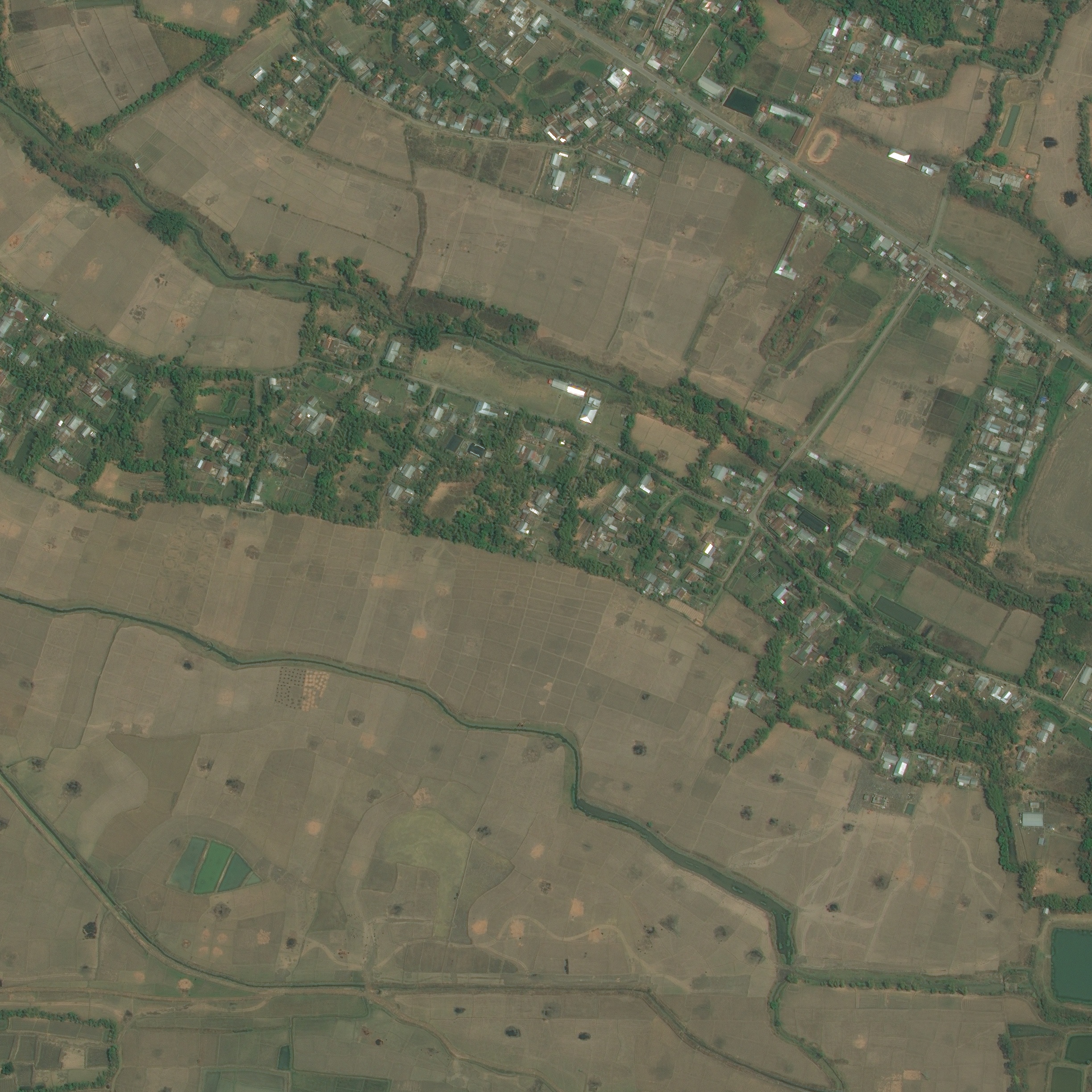}
\end{subfigure}%
\begin{subfigure}{.2\linewidth}
    \centering
    \includegraphics[width=0.7\linewidth]{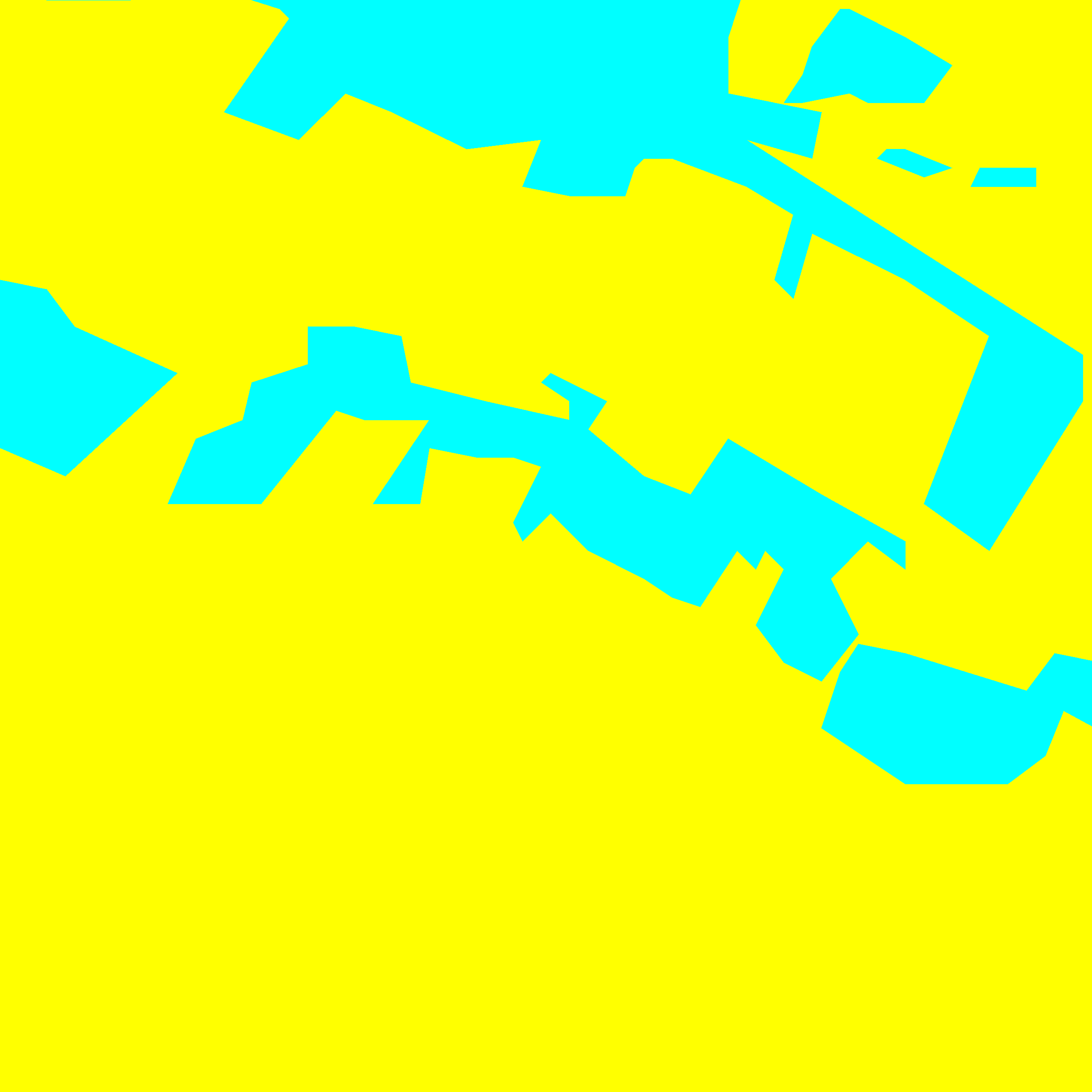}
\end{subfigure}
\begin{subfigure}{.2\linewidth}
    \centering
    \includegraphics[width=0.7\linewidth]{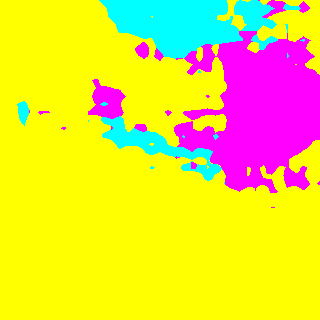}
\end{subfigure}%
\begin{subfigure}{.2\linewidth}
    \centering
    \includegraphics[width=0.7\linewidth]{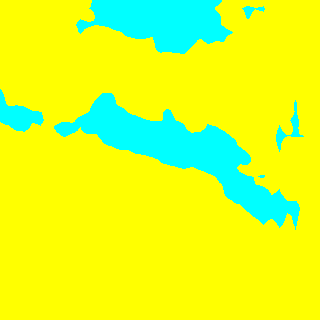}
\end{subfigure}\\
12.5\% 
\begin{subfigure}{.2\linewidth}
    \centering
    \includegraphics[width=0.7\linewidth]{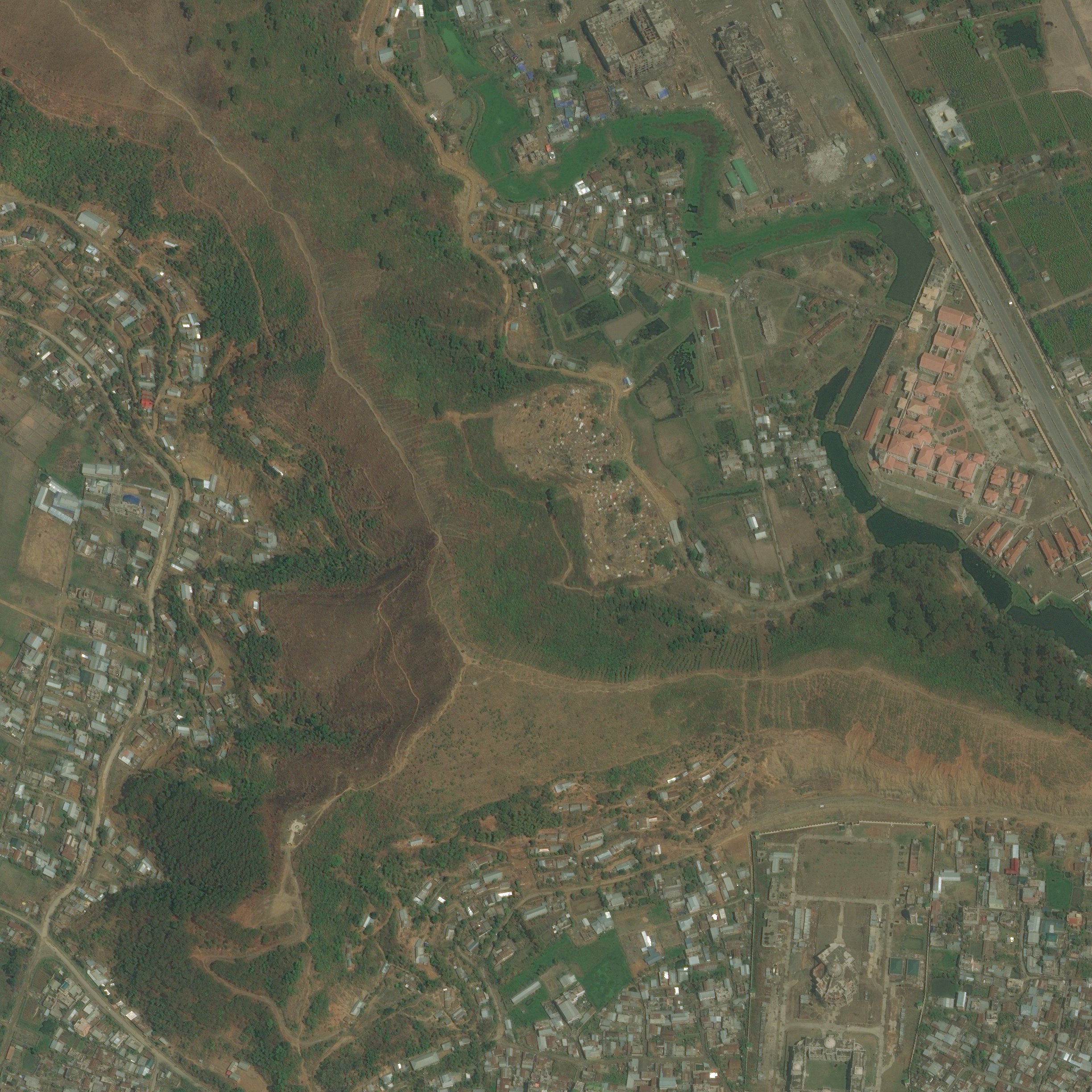}
\end{subfigure}%
\begin{subfigure}{.2\linewidth}
    \centering
    \includegraphics[width=0.7\linewidth]{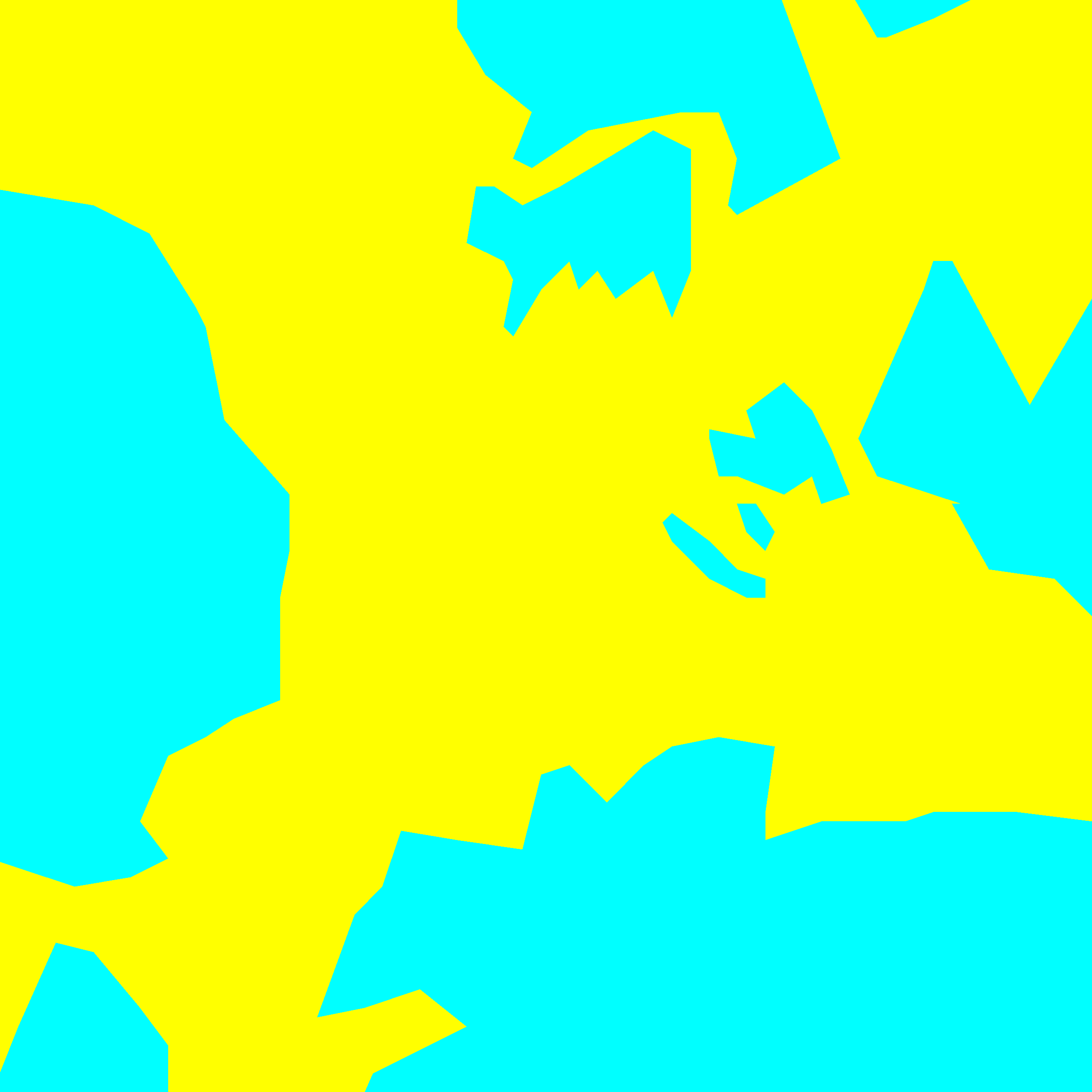}
\end{subfigure}
\begin{subfigure}{.2\linewidth}
    \centering
    \includegraphics[width=0.7\linewidth]{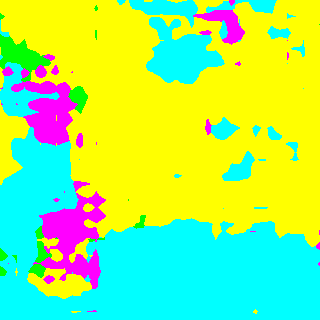}
\end{subfigure}%
\begin{subfigure}{.2\linewidth}
    \centering
    \includegraphics[width=0.7\linewidth]{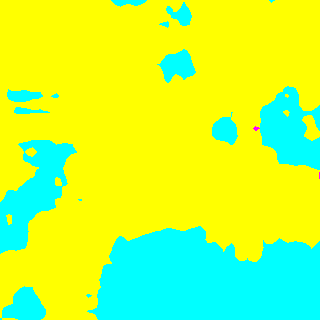}
\end{subfigure}\\
\begin{center}
    \hspace{5pt} a) \textbf{Original Image} \hspace{5pt} b) \textbf{Ground Truth} \hspace{30pt} c) \textbf{Baseline} \hspace{35pt} d) \textbf{Our Results}
\end{center}
\caption{Qualitative Results from the DeepGlobe Land Cover Classification Dataset for different labeled ratios}
\label{qualtitative-deepglobe}
\end{figure*}
 \thispagestyle{empty} 

\begin{algorithm}
\SetAlgoLined
\nonl{\textbf{Input:}}\\
Labeled Ratio, R and unlabeled pool, $X_N$\\
\nonl{\textbf{Define:}}\\
 $Active\_Sampler \leftarrow \text{Active Learning module}$ \label{active_learner}\\
 $\tau \leftarrow \text{Confidence Threshold}$\\
 $G \leftarrow \text{Generator network of conditional GAN}$ \\
 $D \leftarrow \text{Discriminator network of conditional GAN}$\\
  Active sampler from Algorithm \ref{AL}: $X_{N_L}, \_ = Active\_Sampler(R, X_N)$ \\ \label{sampler}
 {\textbf{Get\_Label:}}\
 Obtain pixel-wise labels for the images returned by the active sampler: 
 $Y_{N_L}^P = get\_label(X_{N_L})$\\ \label{lookup}
 Images without pixel-wise labels: $X_{N_U} \leftarrow X_N - X_{N_L}$\label{labeled_samples} \\
 \nonl{\textbf{Semi-Supervised Semantic Segmentation:}} \\
 \While{$i<iterations$}{
 $pseudo\_label = \{\}$\\
 \For{$(x_{N_L}, x_{N_U}, y_{N_L}^P)$ in  $(X_{N_L}, X_{N_U}, Y_{N_L}^P)$}{
    $mask = G(x_{N_L}, x_{N_U}, y_{N_L}^P) $\label{Generator}\\
    $confidence = D(mask, y_{N_L}^P)$ \label{discriminator}\\
        
 \uIf{$confidence >  \tau$}{
     $pseudo\_label = pseudo\_label \bigcup mask$
     }
   }\
Augment pseudo-labels to ground-truth and train the Generator model via self-training loss: 
$Y_{N_L}^P = Y_{N_L}^P \bigcup  pseudo\_label$  \label{union}

  }
\caption{Semi-Supervised semantic segmentation using samples obtained from the Active Learner}
\label{AL_2}
\end{algorithm}

 \thispagestyle{empty} 

\section{Experiments and Results}

\subsection{Datasets and Evaluation Metric}

\noindent
\textbf{UC Merced Land Use Classification Dataset}:
The UC Merced Land Use Classification dataset \cite{yang2010bag} has 2100 RGB images of size 256x256 pixels and 0.3m spatial resolution, with image-level annotations for each of the 21 classes. 
We use the pixel-level annotations for the UC Merced dataset made publicly available by Shao et al. \cite{shao2020multilabel} which has 17 classes as proposed in \cite{chaudhuri2017multilabel}. The dataset was randomly split into training and validation sets with 1680 training images (80\%) 420  validation images (20\%).
\newline
\textbf{DeepGlobe Land Cover Classification Dataset}:
The DeepGlobe land cover classification dataset is comprised of DigitalGlobe Vivid+ images of dimensions 2448x2448 pixels and spatial resolution of 0.5 m. There are 803 pixel-wise annotated training images, each with pixel-wise label covering
seven land cover classes.
Since there are no image-level annotations available for the DeepGlobe dataset, to generate image-level annotations,  we calculate which class contains the highest number of pixels for every image and assign that particular coarse class to the image.  The dataset was randomly split into training and validation sets such that 642 training images (80\%)  and 161 validation images (20\%).
\newline
\textbf{Evaluation Metric: } We use mean Intersection-over-Union (mIoU) as our evaluation metric. 

\subsection{Implementation Details}

\noindent
\textbf{Active Learning for Image Classification}
We used ResNet 101 and ResNet 50 \cite{he2016identity} as our image classification networks for the UC Merced and the DeepGlobe Datasets respectively which were trained using the Cross-Entropy loss. The network was trained using the SGD optimizer with a base learning rate of 0.001 and momentum of 0.9. We used a step learning rate scheduler where the learning rate is dropped by a factor of 0.1 every 7 epochs. We used a batch size of 4 and trained for 50 epochs after each query. Through cross-validation, we found the optimal value of $\alpha_{init}$ = 0.1 and $\beta_Q$ = 0.5. 
We implemented the network using the skorch \cite{skorch} framework. 
The different active learning query strategies were implemented using the modAL toolbox \cite{modAL2018} and trained on a NVIDIA GTX-2080ti GPU.

\noindent
\textbf{Semi-Supervised Semantic Segmentation}
We use a GAN-based semi-supervised semantic segmentation technique proposed by Mittal et al. \cite{mittal2019semi}. The generator is comprised of a segmentation network which in our case is DeepLabv2 \cite{chen2017deeplab} trained with a ResNet-101 \cite{he2016identity} backbone pretrained on the ImageNet dataset \cite{deng2009imagenet}. The discriminator is a binary classifier with four convolutional layers with 4x4 kernels with {64, 128, 256, 512} channels each followed by a Leaky ReLU activation \cite{xu2015empirical} with negative slope of 0.2 and a dropout layer \cite{srivastava2014dropout} with dropout probability of 0.5. 
The segmentation network in the generator is trained with SGD optimizer base learning rate of 2.5e-4, momentum of 0.9, and a weight decay of 5e-4 as described in \cite{hung2018adversarial, mittal2019semi}. The image classification network in the discriminator is trained using the Adam optimizer \cite{kingma2014adam} with a base learning rate of 1e-4.
Through cross-validation, we found the optimal loss weights to be $\lambda_{fm} = 0.1$ and $\lambda_{st} = 1.0$ and
the optimal value of $\tau$ to be 0.6. For the DeepGlobe dataset, we resize each image to 320x320 pixels to reduce the training time. We implemented the network using PyTorch \cite{paszke2017automatic} on NVIDIA Tesla V100 GPUs.

\begin{table}[t] 
    \centering
    \begin{tabular}{p{0.32\linewidth} p{0.15\linewidth} p{0.15\linewidth} p{0.15\linewidth} p{0.0\linewidth}} 
    \hline
      \textbf{Labeled Ratio(R) }  &  \textbf{2\%} & \textbf{5\%} & \textbf{12.5\%} \\
      \hline
       s4GAN \cite{mittal2019semi} (Baseline) & 0.347 $\pm$ 0.0077 & 0.408 $\pm$ 0.0035 & 0.473 $\pm$ 0.0080\\
     
       s4GAN + Entropy (Ours) & \textbf{0.398} & 0.456 & 0.494\\
       
       s4GAN + Margin (Ours) & 0.360 & \textbf{0.480} & \textbf{0.546}\\
       \hline
    \end{tabular}
    \caption{mIoU Scores for the UC Merced Land Use Classification Dataset \cite{yang2010bag}}
\label{tab:ucm_res}
\end{table}
\subsection{Results and Analysis}
 Our baseline is a vanilla s4GAN \cite{mittal2019semi} network, where the labeled data is selected randomly from a given dataset. 
 We report the mean and standard deviation of mIoUs across three experiments with different random seeds for robust evaluation for our baseline method. We compare our approach of using active learning to select representative labeled examples with this baseline. We experiment with two pool-based query strategies, \textit{entropy} and \textit{margin} sampling and demonstrate qualitative and quantitative performance improvements on two datasets, DeepGlobe Land Cover Classification Dataset \cite{demir2018deepglobe}, and UC Merced Land Use Classification Dataset \cite{shao2020multilabel, yang2010bag} over the stated baseline. We evaluated our approach with labeled ratios of 2\%, 5\%, and 12.5\%. The qualitative results in Figures \ref{qualtitative-ucm} and \ref{qualtitative-deepglobe} are shown for the best out of the two sampling strategies for each labeled ratio. Table \ref{data-table} shows the number of labeled images in each dataset for different labeled ratios.
\par
\noindent
\textbf{UC Merced Land Use Classification Dataset:}
Table \ref{tab:ucm_res} shows a quantitative comparison of our method with the baseline for the UC Merced Land Use Classification Dataset \cite{shao2020multilabel, yang2010bag}.
We compare the performance of entropy and margin sample selection strategies with the baseline and show significant and consistent performance improvements. Both the active learning strategies out-perform the baseline by a significant margin. We report a maximum mIoU improvement of close to 15\% with as little as 2\% labeled data, a maximum improvement of about 18\% over the baseline when training with 5\% and 12.5\% labeled data across the two active learning strategies.
 \thispagestyle{empty} 
Figure \ref{qualtitative-ucm} shows how our proposed method qualitatively improves over the UC Merced Dataset baseline for different labeled ratios. Our method predicts a finer coastline with no false positives, even with only as few as 34 labeled images which are 2\% of labeled data (Row 1 of Figure \ref{qualtitative-ucm}).  \thispagestyle{empty} 
Similarly, we demonstrate that even when using only 5\% (85 images) of labeled data (Row 2 of Figure \ref{qualtitative-ucm}).  our method predicts the green river that is camouflaging into the background while the baseline method completely misses it (Row 2 of Figure \ref{qualtitative-ucm}). This shows the importance of having a representative pool of labeled data, especially in a low data regime, as is our case.  With 12.5\% (211 images) of labeled data (Row 3 of Figure \ref{qualtitative-ucm}), our method accurately predicts the complex shape of the airplane (Column d), as opposed to the baseline (Column c), which was confused between multiple unrelated classes. 
\par
\noindent
\textbf{DeepGlobe Land Use Classification Dataset:}  \thispagestyle{empty}  Table \ref{tab:DeepGlobe_res} shows a quantitative comparison of our method with the baseline for the DeepGlobe Land Cover Classification Dataset \cite{demir2018deepglobe}. We report significant performance improvements over the baseline using both entropy and margin sampling strategies. We report a maximum mIoU improvement of close to 27\% with as little as 2\% labeled data, a maximum improvement of about 6\% over the baseline when training with 5\% labeled data, and an improvement of approximately 8\% with 12.5\% labeled data across the two active learning strategies. Figure \ref{qualtitative-deepglobe} shows some visualizations from the DeepGlobe Dataset where it is seen that our method results in fewer false positives than the baseline. 

\begin{table}[t] 
    \centering
    \begin{tabular}{p{0.32\linewidth} p{0.15\linewidth} p{0.15\linewidth} p{0.15\linewidth} p{0.0\linewidth}} 
    \hline
      \textbf{Labeled Ratio(R) }  &  \textbf{2\%} & \textbf{5\%} & \textbf{12.5\%} \\
      \hline
       s4GAN \cite{mittal2019semi} (Baseline) & 0.403 $\pm$ 0.035  & 0.486 $\pm$ 0.018 & 0.511 $\pm$ 0.007\\
       
       s4GAN + Entropy (Ours) & 0.469 & \textbf{0.513}  & \textbf{0.554}\\
       
       s4GAN + Margin (Ours) & \textbf{0.511} & \textbf{0.513} & 0.529\\
       \hline
    \end{tabular}
    \caption{mIoU Scores for the DeepGlobe Land Cover Classification Dataset \cite{demir2018deepglobe}}
\label{tab:DeepGlobe_res}
\end{table}

\section{Conclusion}

This work proposes a method to leverage active learning-based sampling techniques to improve performance on the downstream task of semi-supervised semantic segmentation for land cover classification in satellite images. We do this by intelligently selecting samples for which pixel-wise labels should be obtained using coarse image classification-based active-learning strategies. Our method helps the semi-supervised semantic segmentation network start with an optimal set of labeled examples to help it get the right amount of initial information to learn the suitable representation. We prototype this method for a GAN-based semi-supervised semantic segmentation network, where the labeled images were selected using pool-based active learning strategies.  We demonstrate the efficacy of our method for two satellite image datasets, both quantitatively and qualitatively, and report sizable performance gains. 

\section*{Acknowledgements}
\noindent
We would like to thank our peers who helped us improve our paper with their feedback, in no particular order - Joseph Weber, Wencheng Wu, Gowdhaman Sadhasivam, Surya Teja, Rheeya Uppal, Julius Simonelli, Jing Tian, Kaushik Patnaik.

 \thispagestyle{empty} 

{\small
 \thispagestyle{empty} 
\bibliographystyle{ieee_fullname}
\bibliography{arxiv_submission.bbl}
 \thispagestyle{empty} 
}
\clearpage

\appendix
\section*{Appendices}
\addcontentsline{toc}{section}{Appendices}
\renewcommand{\thesubsection}{\Alph{subsection}}
\setcounter{subsection}{0}

\subsection{Ablation Study}

\subsubsection{Active Learning Parameters}

Tables \ref{tab:UCM_res} and \ref{tab:DeepGlobe_res} show the results of our experiments with different combinations of $\alpha_{init}$ and $\beta_Q$ on UC Merced Land Use Classification \cite{yang2010bag} and the DeepGlobe Land Cover Classification \cite{demir2018deepglobe} datasets respectively. We vary both the parameters between 0.1 and 0.9 for both entropy and margin-based sampling strategies for three different labeled ratios. We found the best performing $\alpha_{init}$ and $\beta_Q$ values to be 0.1 and 0.5 respectively. Overall we noticed out method to be sensitive to changes in $\alpha_{init}$ and $\beta_Q$ as the average difference in the worst performing and best performing model across all labeled ratios and sampling techniques is 4 mIoU points for the UC Merced Land Use Classification Dataset and 2.4 mIoU points for the DeepGlobe Land Cover Classification Dataset. 

\begin{table*}[t]
    \centering
\begin{tabular}{cccccccc}
\hline
\multicolumn{2}{c}{\textbf{Active Learning Parameters}} & \multicolumn{2}{c}{\textbf{2\%}} & \multicolumn{2}{c}{\textbf{5\%}} & \multicolumn{2}{c}{\textbf{12.5\%}}   \\

\textbf{$\alpha_{init}$ }& \textbf{$\beta_Q$}                                   & \textbf{Entropy} & \textbf{Margin}       & \textbf{Entropy} & \textbf{Margin}       & \textbf{Entropy} & \textbf{Margin}             \\
\hline
0.1   & 0.1                                    & 0.381   & 0.358        & 0.423   & 0.450        & 0.484   & 0.497              \\

\textbf{0.1}   & \textbf{0.5}                                    & \textbf{0.398}   & \textbf{0.36}         & \textbf{0.456}   & \textbf{0.48}         & \textbf{0.494}   & \textbf{0.546}              \\

0.9   & 0.9                                    & 0.352   & 0.353        & 0.411   & 0.421        & 0.478   & 0.478              \\
\hline
\end{tabular}
        
    \caption{Ablation Study for the different Active Learning parameters on the UC Merced Land Use Classification Dataset \cite{yang2010bag}}
\label{tab:UCM_res}
\end{table*}

\begin{table*}[t]
    \centering
\begin{tabular}{cccccccc}
\hline
\multicolumn{2}{c}{\textbf{Active Learning Parameters}} & \multicolumn{2}{c}{\textbf{2\%}} & \multicolumn{2}{c}{\textbf{5\%}} & \multicolumn{2}{c}{\textbf{12.5\%}}   \\

\textbf{$\alpha_{init}$} & \textbf{$\beta_{Q}$}         & \textbf{Entropy} & \textbf{Margin}       & \textbf{Entropy} & \textbf{Margin}       & \textbf{Entropy} & \textbf{Margin}             \\
\hline
0.1   & 0.1          & 0.464   & 0.497        & 0.507   & 0.502        & 0.549   & 0.513              \\

\textbf{0.1}   & \textbf{0.5}          & \textbf{0.469}   & \textbf{0.511 }       & \textbf{0.513}   & \textbf{0.513}        & \textbf{0.554}   & \textbf{0.529}              \\

0.9   & 0.9          & 0.449   & 0.462        & 0.495   & 0.498        & 0.527   & 0.512              \\
\hline
\end{tabular}

    \caption{Ablation Study for the different Active Learning parameters on the DeepGlobe Land Cover Classification Dataset \cite{demir2018deepglobe}}
    \label{tab:DeepGlobe_res}
\end{table*}

\begin{table*}[]
    \centering
\begin{tabular}{ccccccc}
\hline
  & \multicolumn{2}{c}{\textbf{2\%}} & \multicolumn{2}{c}{\textbf{5\%}} & \multicolumn{2}{c}{\textbf{12.5\%}}   \\

  \textbf{Backbone}          & \textbf{Entropy} & \textbf{Margin}         & \textbf{Entropy} & \textbf{Margin }        & \textbf{Entropy} & \textbf{Margin}            \\
\hline
VGG-16     & 0.355   & 0.351          & 0.421   & 0.426          & 0.481   & 0.498             \\

Resnet-50  & 0.371   & 0.354          & 0.434   & 0.452          & 0.489   & 0.524             \\

\textbf{Resnet-101} & \textbf{0.398}   & \textbf{0.36}           & \textbf{0.456}   & \textbf{0.48}           & \textbf{0.494}   & \textbf{0.546}             \\
\hline
\end{tabular}
    \caption{Impact of different network architectures for the active learner in UC Merced Land Use Classification Dataset \cite{yang2010bag} on mIoU values}
        \label{tab:UCM_backbone}
\end{table*}

\begin{table*}[]
    \centering
   \begin{tabular}{ccccccc}
   \hline
   & \multicolumn{2}{c}{\textbf{2\%}} & \multicolumn{2}{c}{\textbf{5\%}} & \multicolumn{2}{c}{\textbf{12.5\%}}   \\

 \textbf{Backbone}          & \textbf{Entropy} & \textbf{Margin}         & \textbf{Entropy} & \textbf{Margin}         & \textbf{Entropy} & \textbf{Margin}            \\
\hline
VGG-16     & 0.421   & 0.445          & 0.492   & 0.499          & 0.523   & 0.52              \\

\textbf{Resnet-50}  & \textbf{0.469}   & \textbf{0.511}          & \textbf{0.513}   & \textbf{0.513}          & \textbf{0.554}   & \textbf{0.529 }            \\

Resnet-101 & 0.443   & 0.482          & 0.505   & 0.492          & 0.534   & 0.518             \\
\hline
\end{tabular}
    \caption{Impact of different network architectures for the active learner in the DeepGlobe Land Cover Classification Dataset \cite{demir2018deepglobe} on mIoU values}
        \label{tab:DeepGlobe_backbone}
\end{table*}

\begin{figure}[ht]
    \begin{subfigure} {\linewidth}
        \centering
         \includegraphics[scale=0.5]{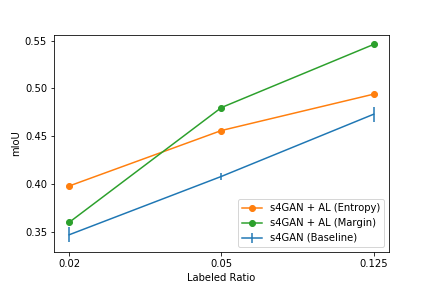}\\
    a)        
    \end{subfigure}\\
    \begin{subfigure}{\linewidth}
        \centering
         \includegraphics[scale=0.5]{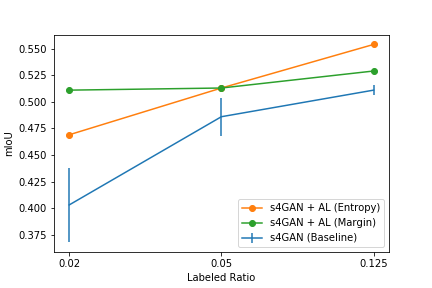} \\
         b)
    \end{subfigure}\\

    \caption{Visualization of quantitative results for different labeled ratios for the a) UC Merced Land Use Classification Dataset \cite{yang2010bag} and, b) DeepGlobe Land Cover Classification Dataset \cite{demir2018deepglobe}}
    \label{fig:result_viz}
\end{figure}

\subsubsection{Network Capacity of Active Learner}

Tables \ref{tab:UCM_backbone} and \ref{tab:DeepGlobe_backbone} show the results of our experiments with different backbone networks on UC Merced Land Use Classification \cite{yang2010bag} and the DeepGlobe Land Cover Classification \cite{demir2018deepglobe} datasets respectively. We experiment with VGG-16 \cite{simonyan2015deep}, ResNet-50 \cite{he2016identity} and ResNet-101 \cite{he2016identity} which have different network capacities. 
We found the best performing backbone network to be ResNet-101 for the UC Merced Land Use Classification dataset and ResNet-50 for the DeepGlobe Land Cover Classification dataset. As shown by the results, the image classification network's capacity for the $learner$ is crucial in determining the quality of the selected samples. Any network with low capacity with respect to the size of the dataset and the number of classes tends to underfit, while any network with a higher capacity than required could overfit and detrimentally affect the downstream task's performance. We noticed our method to be sensitive to networks with different capacities as the average difference in the worst performing and best performing model across all labeled ratios and sampling techniques is 3.3 mIoU points for the UC Merced Land Use Classification Dataset and 2.9 mIoU points for the DeepGlobe Land Cover Classification Dataset. Notably, we see that in most cases, VGG-16 performed significantly performed poorly across all labeled ratios in both the datasets as compared to the ResNet-50 and ResNet-101 models reinforcing the hypothesis that models with insufficient network capacity underperform at the downstream task.

\subsection{Quantitative Evaluation of Diversity}
In this paper, we proposed a method which aims to select the most diverse and representative set of samples to serve as an initial labeled set of data for the semi-supervised network. We empirically showed the success of the proposed method on different datasets. 
In this section, we evaluate the robustness of our method using statistical indices which measure the  diversity of the selected samples.  To achieve this, we choose two diversity indices which are frequently used in ecological studies that measure species diversity, but the same analysis can also be applied to measure diversity of any set of random samples.

\subsubsection{Shannon's Diversity Index}
\label{shannon}

The Shannon index \cite{shannon1948mathematical} was developed from information theory and is based on measuring uncertainty. Shannon's index accounts for both abundance and evenness of the samples present. Shannon index is defined in Equation \ref{shannon-formula}:
\begin{align}
    H(x) = -\sum_{i=1}^N p_i \log p_i
    \label{shannon-formula}
\end{align}
 In our case, each sample is a pixel. Hence, $p_i$ indicates the probability that a given pixel belongs to class $i$. N indicates the total number of classes that a given pixel can belong to. 
Thus, we are measuring how diverse are the samples selected by the active learning method as compared to samples selected randomly. Therefore, samples with a large number of pixels from different classes that are evenly distributed are the most diverse. On the other hand, samples that are dominated by pixels from one class are the least diverse. We report the value of Shannon diversity index for our baseline method averaged across our three experiments with different random seeds and for samples selected by both the active learning techniques.
 Intuitively, Shannon's index quantifies the uncertainty in predicting the class to which a given pixel belongs and hence a higher value of Shannon diversity index indicates a more diverse set of samples.
 \par
 Our results for Shannon's diversity index are shown in Tables \ref{tab:UCM_Shannon} and \ref{tab:Deepglobe_Shannon} for the UC Merced Land Use Classification \cite{yang2010bag} and DeepGlobe Land Cover Classification \cite{demir2018deepglobe} datasets respectively. We notice  a strong correlation between the mIoU values reported in the paper for the baseline and active learning strategies and the values of the Shannon's diversity index obtained for the respective experiments. 

\begin{table}[ht] 
    \centering
    \begin{tabular}{p{0.4\linewidth}p{0.15\linewidth} p{0.15\linewidth}p{0.15\linewidth}p{0.0\linewidth}} 
    \hline
      \textbf{Labeled Ratio(R) }  &  \textbf{2\%} & \textbf{5\%} & \textbf{12.5\%} \\
      \hline
       s4GAN \cite{mittal2019semi} (Baseline) & 1.96 $\pm$ 0.08   & 2.16  $\pm$ 0.02 & 2.14 $\pm$ 0.03 \\
       
       s4GAN + Entropy (Ours) & 2.10 & 2.20  & 2.22\\
       
       s4GAN + Margin (Ours) & 2.08 & 2.22  & 2.25 \\
       \hline
    \end{tabular}
    \caption{Shannon's Diversity Index for the UC Merced Land Use Classification Dataset \cite{yang2010bag} (Higher the better)}
\label{tab:UCM_Shannon}
\end{table}

\begin{table}[ht] 
    \centering
    \begin{tabular}{p{0.4\linewidth}p{0.15\linewidth}p{0.15\linewidth}p{0.15\linewidth}p{0.0\linewidth}} 
    \hline
      \textbf{Labeled Ratio(R) }  &  \textbf{2\%} & \textbf{5\%} & \textbf{12.5\%} \\
      \hline
       s4GAN \cite{mittal2019semi} (Baseline) & 1.01 $\pm$ 0.04   & 1.16  $\pm$ 0.05 & 1.19 $\pm$ 0.14 \\
       
       s4GAN + Entropy (Ours) & 1.06 & 1.25  & 1.38\\
       
       s4GAN + Margin (Ours) & 1.09 & 1.24  & 1.36 \\
       \hline
    \end{tabular}
    \caption{Shannon's Diversity Index for the DeepGlobe Land Cover Classification Dataset \cite{demir2018deepglobe} (Higher the better)}
\label{tab:Deepglobe_Shannon}
\end{table}

\subsubsection{Simpson's Diversity Index}
Traditionally, Simpson's Diversity Index \cite{simpson1949medicion}
measures the probability that two individuals randomly selected from a sample will belong to the same species (or some category other than species). We extend it to our use case to measure the diversity of the selected samples. To make it easier and intuitive to understand the relevance of this index, we use the inverse Simpson index. Thus,  greater the value, the greater the sample diversity.  In this case, the index represents the probability that two individuals randomly selected from a sample will belong to different species. Thus, the inverse Simpson index is defined in \ref{simpson-formula}:
\begin{align}
D = 1-\frac{\sum(n(n-1))}{N(N-1)}
\label{simpson-formula}
\end{align}
\newline
where, \newline
$n$ = the number of pixels belonging to class $i$, 
\newline
\noindent
$N$ = total number of classes that exist in the dataset. \par
Similar to Shannon's index in Section \ref{shannon}, we report results on the UC Merced Land Use Classification \cite{yang2010bag} and the DeepGlobe Land Cover Classification \cite{demir2018deepglobe} datasets in Tables  \ref{tab:UCM_Simpson} and \ref{tab:Deepglobe_Simpson}
We show that both the active learning sampling strategies used in this paper yield more diverse set of samples and show strong correlation with the mIoU values reported on these datasets in the paper.

\begin{table}[] 
    \centering
    \begin{tabular}{p{0.4\linewidth}p{0.15\linewidth}p{0.15\linewidth}p{0.15\linewidth}p{0.0\linewidth}} 
    \hline
      \textbf{Labeled Ratio(R) }  &  \textbf{2\%} & \textbf{5\%} & \textbf{12.5\%} \\
      \hline
       s4GAN \cite{mittal2019semi} (Baseline) & 0.79 $\pm$ 0.03   & 0.83  $\pm$ 0.009 & 0.83 $\pm$ 0.008 \\
       
       s4GAN + Entropy (Ours) & 0.85 & 0.84  & 0.85\\
       
       s4GAN + Margin (Ours) & 0.82 & 0.86  & 0.87 \\
       \hline
    \end{tabular}
    \caption{Simpson's Diversity Index for the UC Merced Land Use Classification Dataset \cite{yang2010bag} (Higher the better)}
\label{tab:UCM_Simpson}
\end{table}

\begin{table}[] 
    \centering
    \begin{tabular}{p{0.4\linewidth}p{0.15\linewidth}p{0.15\linewidth}p{0.15\linewidth}p{0.0\linewidth}} 
    \hline
      \textbf{Labeled Ratio(R) }  &  \textbf{2\%} & \textbf{5\%} & \textbf{12.5\%} \\
      \hline
       s4GAN \cite{mittal2019semi}  (Baseline) & 0.55 $\pm$ 0.04   & 0.64  $\pm$ 0.01 & 0.65 $\pm$ 0.02 \\
       
       s4GAN + Entropy (Ours) & 0.58 & 0.73  & 0.71\\
       
       s4GAN + Margin (Ours) & 0.62 & 0.71  & 0.68\\
       \hline
    \end{tabular}
    \caption{Simpson's Diversity Index for the DeepGlobe Land Cover Classification Dataset \cite{demir2018deepglobe} (Higher the better)}
\label{tab:Deepglobe_Simpson}
\end{table}

\begin{figure*}[ht]
    \centering
    \begin{subfigure}{0.45\linewidth}
        \centering
        \includegraphics[scale=0.35]{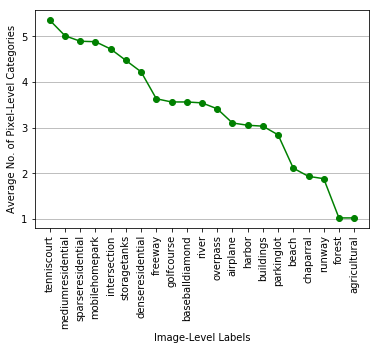}\\
        a)
    \end{subfigure}
        \begin{subfigure}{0.45\linewidth}
        \centering
        \includegraphics[scale=0.35]{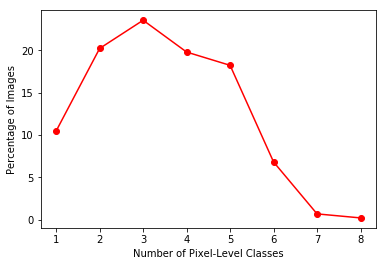}\\
        b)
    \end{subfigure}\\
        \centering
    \begin{subfigure}{0.45\linewidth}
        \centering
        \includegraphics[scale=0.35]{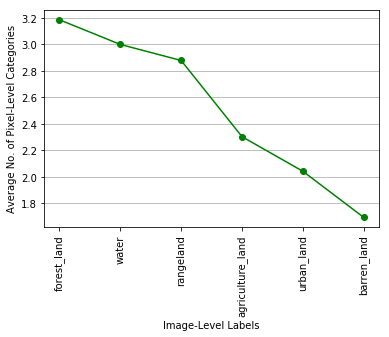}\\
        c)
    \end{subfigure}
        \begin{subfigure}{0.45\linewidth}
        \centering
        \includegraphics[scale=0.35]{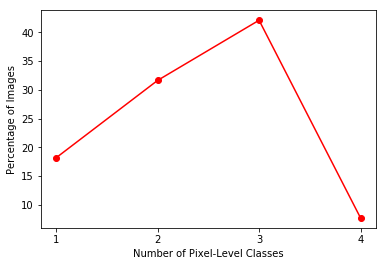}\\
        d)
    \end{subfigure}
    \caption{(a) Average number of semantic categories per image-level category for UC Merced Land Use Classification Dataset. (b) Percentage of images containing vs. number of pixel-level categories per image for UC Merced Land Use Classification Dataset. (c) Average number of semantic categories per image-level category for DeepGlobe Land Cover Classification Dataset. (d) Percentage of images containing vs. number of pixel-level categories per image for DeepGlobe Land Cover Classification Dataset}
    \label{fig:data_fig}
\end{figure*}

\subsection{Discussion}

\subsubsection{Applicability of Our Method to Land Use Classification}

The average number of semantic categories per scene in the UC Merced and DeepGlobe Landuse Classification datasets used in this paper is 3.39 and 2.51 respectively as depicted by figure \ref{fig:data_fig}. This implies that a given scene from the UCM dataset with a given image-level label will have 3 or more different pixel-level labels(semantic categories). Similarly, for the DeepGlobe dataset, we have about 2 or more semantic categories per scene on an average. UCM dataset has a total of 18 semantic categories and DeepGlobe has 6 semantic categories. Thus, each satellite scene in the UCM dataset has about 18\% of all pixel level labels and similarly each satellite scene in the DeepGlobe dataset has about 42\% of all pixel-level labels on an average.  Figure \ref{fig:data_fig} also shows us that about 90\% of scenes in the UCM dataset have more than 1 semantic category and similarly about 80\% of scenes in the DeepGlobe dataset have more than 1 semantic category. This number if quite high when we compare this statistics with that in some generic standard dataset. For instance, consider the COCO dataset \cite{lin2014microsoft}. Less that 30\% of the images in the COCO dataset have more than 1 semantic category. This tells us that the landuse scenes in the domain of satellite imagery are inherently more diverse and hence our method is highly applicable specifically for land use classification  in satellite images. We will get a more diverse set of samples for satellite domain as compared to using our method on generic datasets like COCO.

\subsubsection{Suitability of s4GAN as our baseline}
\cite{mittal2019semi} propose to fuse the output of the s4GAN network with another image classification-based network called MLMT \cite{tarvainen2017mean} during inference to reduce false positives. This MLMT branch uses an image classification network to output a confidence score for every category in the dataset. This output is combined with the pixel level output of the s4GAN network to reduce the number of false positives in the segmentation network. Therefore, one major constraint for using MLMT is that there should be a one-to-one correspondence between the image-level and the pixel-level labels. This would mean that the number of image-level categories should equal the number of pixel-level categories in a dataset. However, this does not always hold in the case of land use classification. An image-level label for land use classification in a satellite scene indicates predominant usage of land. However, the same scene can have multiple semantic categories. This prevents us from using MLMT as done by \cite{mittal2019semi} as our baseline for the task of land use classification.

\begin{figure*}[ht]
\centering
\begin{subfigure}{.18\linewidth}
    \centering
    \includegraphics[width=0.7\linewidth]{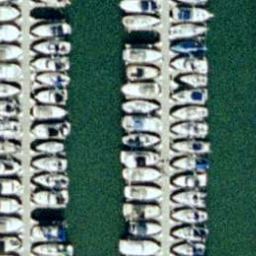}
\end{subfigure}%
\begin{subfigure}{.18\linewidth}
    \centering
    \includegraphics[width=0.7\linewidth]{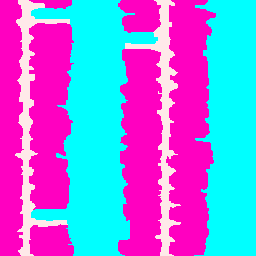}
\end{subfigure}
\begin{subfigure}{.18\linewidth}
    \centering
    \includegraphics[width=0.7\linewidth]{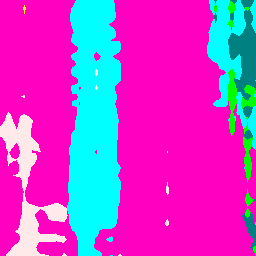}
\end{subfigure}%
\begin{subfigure}{.18\linewidth}
    \centering
    \includegraphics[width=0.7\linewidth]{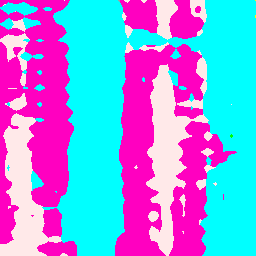}
\end{subfigure}\\
\begin{subfigure}{.18\linewidth}
    \centering
    \includegraphics[width=0.7\linewidth]{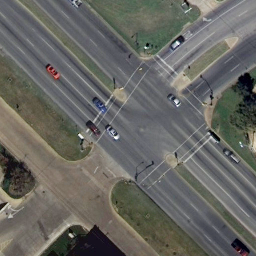}
\end{subfigure}%
\begin{subfigure}{.18\linewidth}
    \centering
    \includegraphics[width=0.7\linewidth]{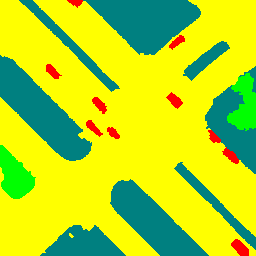}
\end{subfigure}
\begin{subfigure}{.18\linewidth}
    \centering
    \includegraphics[width=0.7\linewidth]{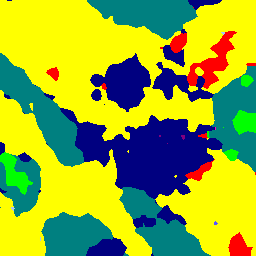}
\end{subfigure}%
\begin{subfigure}{.18\linewidth}
    \centering
    \includegraphics[width=0.7\linewidth]{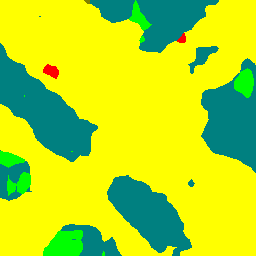}
\end{subfigure}\\
\begin{subfigure}{.18\linewidth}
    \centering
    \includegraphics[width=0.7\linewidth]{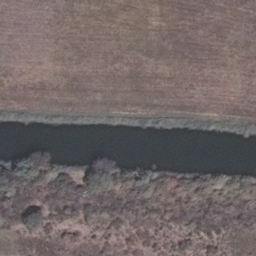}
\end{subfigure}%
\begin{subfigure}{.18\linewidth}
    \centering
    \includegraphics[width=0.7\linewidth]{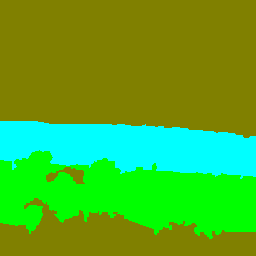}
\end{subfigure}
\begin{subfigure}{.18\linewidth}
    \centering
    \includegraphics[width=0.7\linewidth]{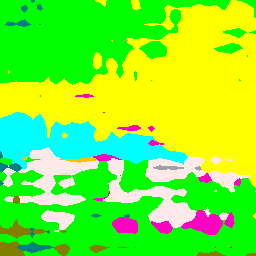}
\end{subfigure}%
\begin{subfigure}{.18\linewidth}
    \centering
    \includegraphics[width=0.7\linewidth]{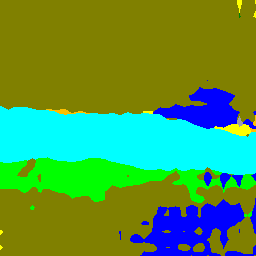}
\end{subfigure}\\
\begin{subfigure}{.18\linewidth}
    \centering
    \includegraphics[width=0.7\linewidth]{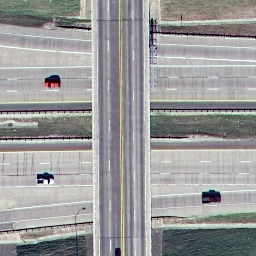}
\end{subfigure}%
\begin{subfigure}{.18\linewidth}
    \centering
    \includegraphics[width=0.7\linewidth]{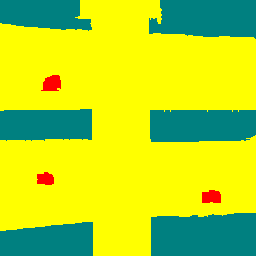}
\end{subfigure}
\begin{subfigure}{.18\linewidth}
    \centering
    \includegraphics[width=0.7\linewidth]{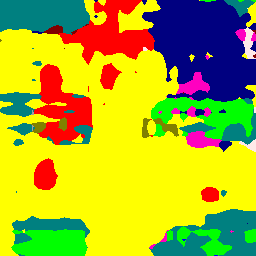}
\end{subfigure}%
\begin{subfigure}{.18\linewidth}
    \centering
    \includegraphics[width=0.7\linewidth]{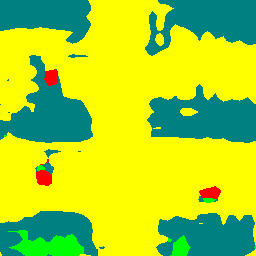}
\end{subfigure}\\
\begin{center}
    \hspace{5pt} a) \textbf{Original Image} \hspace{5pt} b) \textbf{Ground Truth} \hspace{30pt} c) \textbf{Baseline} \hspace{35pt} d) \textbf{Our Results}
\end{center}
\caption{Qualitative Results from the UC Merced Land Use Classification Dataset for 2\% labeled data}
\label{2-qualtitative-ucm}
\end{figure*}

\subsection{More Qualitative Evaluation}

In this section, we provide more qualitative results from our best performing active learning strategies and compare them to our baseline for the UC Merced Land Use Classification Dataset \cite{yang2010bag}. 
\par
Figure \ref{2-qualtitative-ucm} compares the performance of our method with the baseline when trained with 2\%  labeled data. Row 1 shows how our method predicts the row of boats parked on the harbor better than the baseline method. Rows 2, 3, and 4 show that the baseline method gets confused between multiple unrelated classes, whereas our method reasonably predicts the correct classes. 
\par
\begin{figure*}[ht]
\centering
\begin{subfigure}{.18\linewidth}
    \centering
    \includegraphics[width=0.7\linewidth]{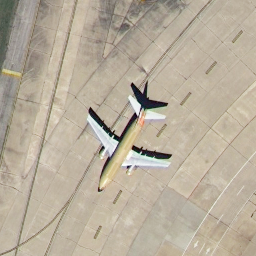}
\end{subfigure}%
\begin{subfigure}{.18\linewidth}
    \centering
    \includegraphics[width=0.7\linewidth]{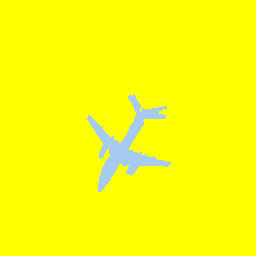}
\end{subfigure}
\begin{subfigure}{.18\linewidth}
    \centering
    \includegraphics[width=0.7\linewidth]{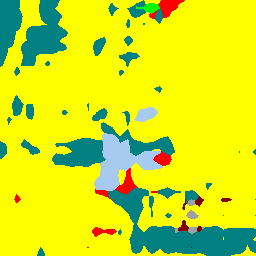}
\end{subfigure}%
\begin{subfigure}{.18\linewidth}
    \centering
    \includegraphics[width=0.7\linewidth]{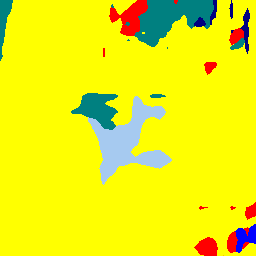}
\end{subfigure}\\
\begin{subfigure}{.18\linewidth}
    \centering
    \includegraphics[width=0.7\linewidth]{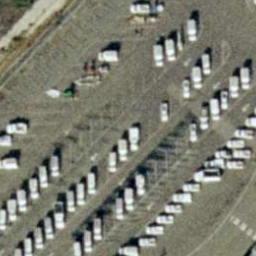}
\end{subfigure}%
\begin{subfigure}{.18\linewidth}
    \centering
    \includegraphics[width=0.7\linewidth]{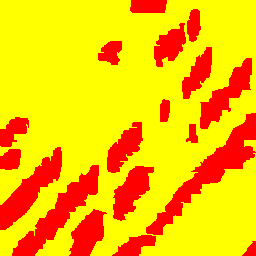}
\end{subfigure}
\begin{subfigure}{.18\linewidth}
    \centering
    \includegraphics[width=0.7\linewidth]{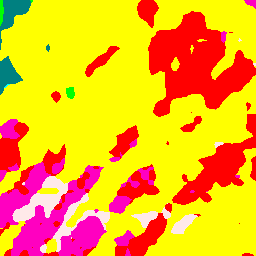}
\end{subfigure}%
\begin{subfigure}{.18\linewidth}
    \centering
    \includegraphics[width=0.7\linewidth]{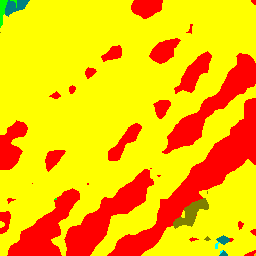}
\end{subfigure}\\
\begin{subfigure}{.18\linewidth}
    \centering
    \includegraphics[width=0.7\linewidth]{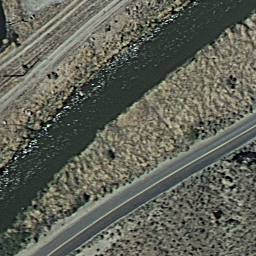}
\end{subfigure}%
\begin{subfigure}{.18\linewidth}
    \centering
    \includegraphics[width=0.7\linewidth]{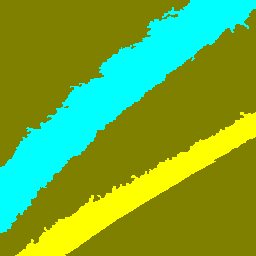}
\end{subfigure}
\begin{subfigure}{.18\linewidth}
    \centering
    \includegraphics[width=0.7\linewidth]{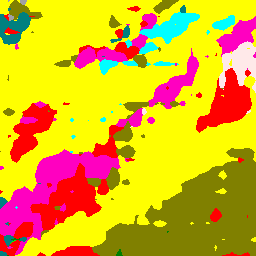}
\end{subfigure}%
\begin{subfigure}{.18\linewidth}
    \centering
    \includegraphics[width=0.7\linewidth]{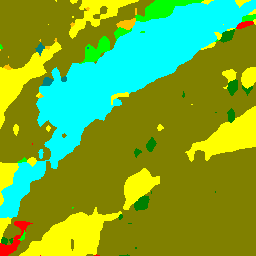}
\end{subfigure}\\
\begin{subfigure}{.18\linewidth}
    \centering
    \includegraphics[width=0.7\linewidth]{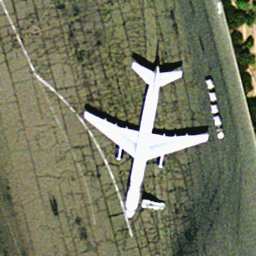}
\end{subfigure}%
\begin{subfigure}{.18\linewidth}
    \centering
    \includegraphics[width=0.7\linewidth]{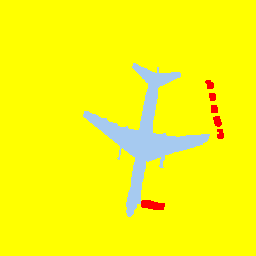}
\end{subfigure}
\begin{subfigure}{.18\linewidth}
    \centering
    \includegraphics[width=0.7\linewidth]{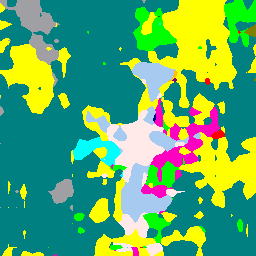}
\end{subfigure}%
\begin{subfigure}{.18\linewidth}
    \centering
    \includegraphics[width=0.7\linewidth]{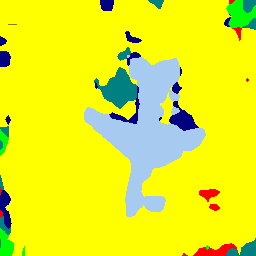}
\end{subfigure}\\
\begin{center}
    \hspace{5pt} a) \textbf{Original Image} \hspace{5pt} b) \textbf{Ground Truth} \hspace{30pt} c) \textbf{Baseline} \hspace{35pt} d) \textbf{Our Results}
\end{center}
\caption{Qualitative Results from the UC Merced Land Use Classification Dataset for 5\% labeled data}
\label{5-qualtitative-ucm}
\end{figure*}

Similarly, Figure \ref{5-qualtitative-ucm} qualitatively compares the performance of our method with the baseline when trained with 5\% labeled data. Rows 1 and 4 show an example of our method predicting the complex shape of airplanes better than the baseline method. Row 2 shows the baseline method being confused between cars in a parking lot and boats parked along a harbor, whereas our method predicts cars parked close together correctly. Row 3 shows how the baseline method completely misses the river and gets confused between multiple classes, while our method predicts the river reasonably well.
\par
\begin{figure*}[ht]
\centering
\begin{subfigure}{.18\linewidth}
    \centering
    \includegraphics[width=0.7\linewidth]{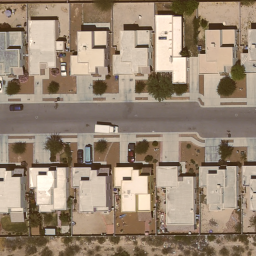}
\end{subfigure}%
\begin{subfigure}{.18\linewidth}
    \centering
    \includegraphics[width=0.7\linewidth]{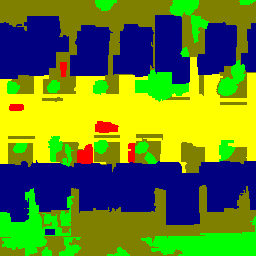}
\end{subfigure}
\begin{subfigure}{.18\linewidth}
    \centering
    \includegraphics[width=0.7\linewidth]{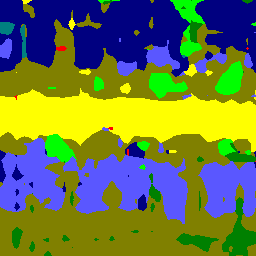}
\end{subfigure}%
\begin{subfigure}{.18\linewidth}
    \centering
    \includegraphics[width=0.7\linewidth]{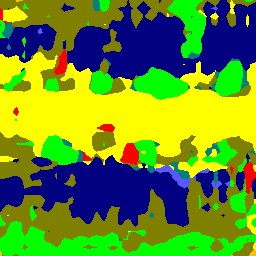}
\end{subfigure}\\
\begin{subfigure}{.18\linewidth}
    \centering
    \includegraphics[width=0.7\linewidth]{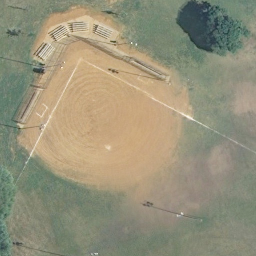}
\end{subfigure}%
\begin{subfigure}{.18\linewidth}
    \centering
    \includegraphics[width=0.7\linewidth]{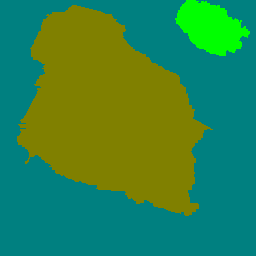}
\end{subfigure}
\begin{subfigure}{.18\linewidth}
    \centering
    \includegraphics[width=0.7\linewidth]{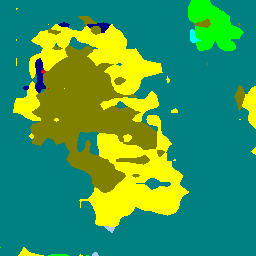}
\end{subfigure}%
\begin{subfigure}{.18\linewidth}
    \centering
    \includegraphics[width=0.7\linewidth]{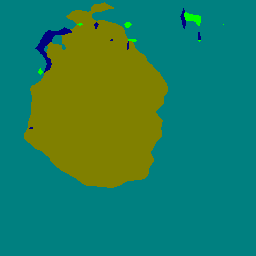}
\end{subfigure}\\
\begin{subfigure}{.18\linewidth}
    \centering
    \includegraphics[width=0.7\linewidth]{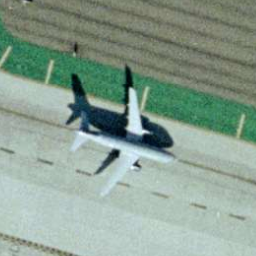}
\end{subfigure}%
\begin{subfigure}{.18\linewidth}
    \centering
    \includegraphics[width=0.7\linewidth]{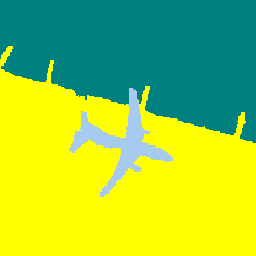}
\end{subfigure}
\begin{subfigure}{.18\linewidth}
    \centering
    \includegraphics[width=0.7\linewidth]{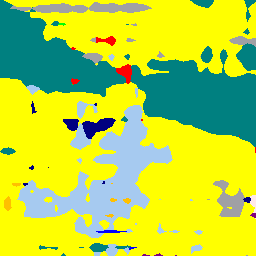}
\end{subfigure}%
\begin{subfigure}{.18\linewidth}
    \centering
    \includegraphics[width=0.7\linewidth]{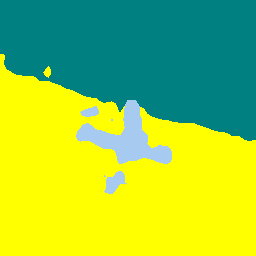}
\end{subfigure}\\
\begin{subfigure}{.18\linewidth}
    \centering
    \includegraphics[width=0.7\linewidth]{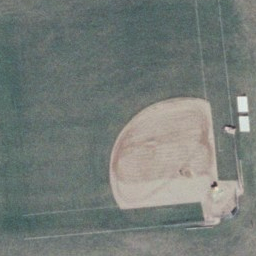}
\end{subfigure}%
\begin{subfigure}{.18\linewidth}
    \centering
    \includegraphics[width=0.7\linewidth]{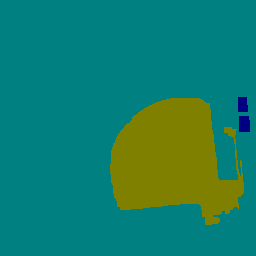}
\end{subfigure}
\begin{subfigure}{.18\linewidth}
    \centering
    \includegraphics[width=0.7\linewidth]{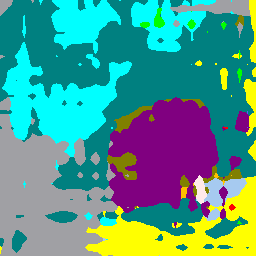}
\end{subfigure}%
\begin{subfigure}{.18\linewidth}
    \centering
    \includegraphics[width=0.7\linewidth]{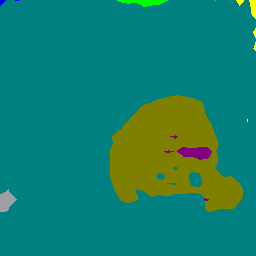}
\end{subfigure}\\
\begin{center}
    \hspace{5pt} a) \textbf{Original Image} \hspace{5pt} b) \textbf{Ground Truth} \hspace{30pt} c) \textbf{Baseline} \hspace{35pt} d) \textbf{Our Results}
\end{center}
\caption{Qualitative Results from the UC Merced Land Use Classification Dataset for 12.5\% labeled data}
\label{12.5-qualtitative-ucm}
\end{figure*}

Finally, Figure \ref{12.5-qualtitative-ucm} shows some qualitative examples of how our method outperforms the baseline when trained with 12.5\% labeled data. Row 1 shows the baseline being confused between buildings and mobile homes, while our method predicts buildings in a dense residential setting more accurately. Rows 2 and 4 show our method predicting the baseball diamond structures accurately without being confused between other classes. Similarly, as shown by Row 3, our method predicts the contours of the airplane better than the baseline.

 \end{document}